\documentclass{article}

\usepackage{arxiv}
\usepackage{hyperref}
\usepackage{helvet} 
\usepackage{graphicx}
\usepackage{cite}
\usepackage{amssymb,amsmath,amsthm}
\usepackage{mathtools}
\usepackage{multirow}
\usepackage{booktabs}	
\usepackage{float}	
\usepackage{amsmath,amsfonts,latexsym,amssymb}
\usepackage[mathscr]{euscript}
\usepackage{mathtools}
\usepackage{subfigure}
\usepackage[ruled,linesnumbered]{algorithm2e}

\usepackage{enumitem}
\usepackage{bm}
\usepackage{color,soul} 
\usepackage{array}
\usepackage[ruled,linesnumbered]{algorithm2e}

\usepackage{siunitx} 
\usepackage{cancel} 
\usepackage{comment} 

\renewcommand{\equationautorefname}{Eq.}

\newcommand{\req}[1]{(\ref{#1})}
\def\equationautorefname~#1\null{%
  Eq.~(#1)\null
  }

\newcommand{\partialsecondderi}[2]{\ensuremath{\displaystyle{\frac {\partial^2 #1} {\partial {#2}^2} } } }

\newcommand{\partialfirstderi}[2]{\ensuremath{\displaystyle{\frac {\partial #1} {\partial #2} } } }

\title{Machine learning based digital twin for dynamical systems with multiple time-scales}

\author{
  Souvik~Chakraborty\\
  Department of Applied Mechanics\\
  Indian Institute of Technology Delhi\\
  New Delhi, India \\
  \texttt{csouvik41@gmail.com} \\
   \And
 Sondipon~Adhikari \\
  School of Engineering\\
  Swansea University\\
  Swansea, SA1 8EN \\
  \texttt{s.adhikari@swansea.ac.uk} \\
}

\begin{document}
\maketitle

\begin{abstract}
Digital twin technology has a huge potential for widespread applications in different industrial sectors such as infrastructure, aerospace, and automotive. However, practical adoptions of this technology have been slower, mainly due to a lack of application-specific details. Here we focus on a digital twin framework for linear single-degree-of-freedom structural dynamic systems evolving in two different operational time scales in addition to its intrinsic dynamic time-scale. Our approach strategically separates into two components -- (a) a physics-based nominal model for data processing and response predictions, and (b) a data-driven machine learning model for the time-evolution of the system parameters. The physics-based nominal model is system-specific and selected based on the problem under consideration. On the other hand, the data-driven machine learning model is generic. For tracking the multi-scale evolution of the system parameters, we propose to exploit a mixture of experts as the data-driven model. Within the mixture of experts model, Gaussian Process (GP) is used as the expert model. The primary idea is to let each expert track the evolution of the system parameters at a single time-scale. For learning the hyperparameters of the `mixture of experts using GP', an efficient framework the exploits expectation-maximization and sequential Monte Carlo sampler is used. Performance of the digital twin is illustrated on a multi-timescale dynamical system with stiffness and/or mass variations. The digital twin is found to be robust and yields reasonably accurate results. One exciting feature of the proposed digital twin is its capability to provide reasonable predictions at future time-steps. Aspects related to the data quality and data quantity are also investigated.
\end{abstract}


\section{Introduction}
\label{sec:intro}
Design and analysis of complex engineering systems using high fidelity computational simulations are an integral part of modern engineering practice. 
In the context of aerospace and mechanical engineering, computational simulations were historically employed to support conceptual design, prototyping, manufacturing, production, 
test-data correlation and safety assessment. Over the last decade, there has been a shift in taking advantage of computational simulations in providing service throughout the whole product life cycle \cite{Rasheed_et_al_2020,tao}, going well beyond the production stage. In the context of civil infrastructure, the idea of fusing digital information with real-life structures are also evolving at a significant pace \cite{Arup_DT_report}. 
The methodologies, algorithms, techniques, software and computer applications which mimic the evolution of a complex real system through computational and digital means are broadly termed as `digital twins'.  The global digital twin technology market size was valued at \$2.26 billion in 2017 and is expected to expand at an exceptional compound annual growth rate (CAGR) of 38.2\% from 2018 to 2025, according to a recent report by Grand View Research Inc \cite{DT_CAGR_report}. Therefore,  the market size is expected to reach a staggering \$26.07 billion by 2025. The key technologies and solutions that enabled the espousal of the technology include Artificial Intelligence (AI)/ Machine Learning (ML), and Internet-of-Things (IoT), among others. 
Factors, such as growing usage of connected devices across various organizations, increasing adoption of cloud platforms, and the emergence of high-speed networking technologies will fuel the growth of the digital twin technology.

By its very definition, digital twins are extremely diverse and can mean very different approaches to different applications. In this paper, we are interested in digital twins of structural dynamic systems, as many physical engineering systems can be expressed in this form.  A digital twin is a virtualized proxy of a real physical dynamic system. While a numerical model of a physical system attempts to closely match the behaviour of a dynamic system, the digital twin also tracks the temporal evolution of the dynamic system. 
Once a digital twin has been trained and developed, it can be used to make crucial decisions at a point in time which is significantly far in the future from the time of manufacturing of an engineering dynamic system. A  general mathematical framework for digital twin has been suggested in \cite{worden}.   More specific approaches to developing digital twins include  prognostics and health monitoring  \cite{Heyns_mssp_2020,Wang_etal_2019,tao2019digital,millwater2019probabilistic,zhou2019digital}, 
manufacturing \cite{debroy,haag,Lu_etal_2020,parkoperation,he2019digital}, automotive and aerospace engineering \cite{li,kapteyn2020toward,tuegel,hoodorozhkov2020digital}, to mention a few. 
These references give an excellent idea of what can be achieved currently using  the digital twin approach.

Dynamic systems differ crucially from other systems due to the fact that their response due to external excitations change with time. The rate of change depends on their characteristics time period. This is a fundamental property of a dynamic system. Typically, smaller structures have faster time periods and larger structures, for example, a large wind turbine, have slower time periods. However, irrespective of their characteristics time periods, the time-scale of their operational life is large. An example, for a large wind turbine, the time period is in the order of 10s of seconds \cite{jp77}, while its operation life is in 10s of years. To address this fundamental mismatch in the time-scales, the digital twin proposed in \cite{Ganguli2020} 
explicitly considered two different time scales. The intrinsic timescale is a fast time scale, while the operational time scale is a slow time scale. 
Physics-based methods, such as the finite element method, are used for dynamic evolution in the intrinsic timescale, while data-based methods (e.g., surrogate models) are used for dynamic evolution in the operational timescale. The digital twin of a complex dynamic system will arise from the fusion of physics and data-based approaches. The separation of computational approaches based on the two different time scales was exploited in \cite{chakraborty2020role} where Gaussian Process Emulators (GPE) was used in the slow time-scale. 

One key aspect of the digital twin technology is to use the sensor data collected from the physical system to update the digital twin and then use the same for predicting the future states. In this regard, the role of machine learning (ML) algorithms become enormous. One of the reasons behind the recent thrust in digital twin technology is the development of advanced ML algorithms (e.g., deep neural network \cite{chakraborty2020simulation,goswami2020transfer}, Gaussian process \cite{Bilionis2013multi, Bilionis2012multi, nayek2019gaussian}) that can be readily used to update the model and make future predictions. For example, \cite{Heyns_mssp_2020} used two deep learning algorithms within the digital twin framework for prognosis and diagnosis of systems. Similarly, in \cite{chakraborty2020role}, GPE was used for learning the evolution of the system parameters. A detailed review of the impact of machine learning algorithms on the digital twin technology can be found in \cite{kaur2020convergence}.

Although the separation of two time-scales provides a logical framework for developing computational and mathematical methods for the construction of digital twins, questions remain on how to define and propose the slow time scale. The fast time-scale is a fundamental property of dynamical systems and therefore is unambiguous for a given system. The same is not true for the slow operational time scale. In \cite{Ganguli2020,chakraborty2020role}  the idea of a single time-scale for the evolution of the digital twin was used for its entire operational period.  However, there is no physical or mathematical reason as to why this must be restricted to only one time-scale. It is perfectly possible that various factors in a complex digital twin evolve at different time scales. For example, the mass of a system can change due to corrosion, while the stiffness of a system can degrade due to fatigue. These two processes will have a very different time scale of evolution. Therefore, in a more general setting, a digital twin can evolve in different time scales in addition to its intrinsic time scale. The key idea proposed and investigated in this paper is that the digital twin of a dynamic system evolves in two different operational time-scales. In principle, there can be more than two operational time-scales. Approaches proposed in the paper can form the basis of considering such problems.  

The rest of the paper is organised as follows. In \autoref{sec:ps}, the problem undertaken in this study is discussed. Details about the proposed digital twin framework for the multiscale dynamical system is discussed in \autoref{sec:dt_ms}. The performance of the proposed digital twin in capturing the (multiscale) temporal evolution of the system parameters is presented in \autoref{sec:illus}. Some key features of the proposed framework and key findings of this study are discussed in \autoref{sec:discussions}. Finally, \autoref{sec:conclusions} presents the concluding remarks.

\section{The problem statement}
\label{sec:ps}
We consider a physical system that can be represented by 
a single degree of freedom (SDOF) spring mass and damper system.
\begin{equation}\label{eq:eom}
    m_0 \frac{\text d^2u_0 \left(t\right)}{\text d t^2} + c_0 \frac{\text d u_0 \left(t\right)}{\text dt} + k_0 u_0\left(t\right) = f_0 \left( t\right),
\end{equation}
where $m_0$, $c_0$ and $k_0$ are, respectively the mass,
damping and stiffness of the system. Here $t$ is the intrinsic time of the system.
Equation \req{eq:eom} is often referred to as the `nominal system' and $m_0$, $c_0$ and $k_0$ as the nominal mass, nominal damping and
nominal stiffness, respectively.
$f_0 \left(t\right)$ and $u_0 \left(t\right)$
are respectively the forcing function and the dynamic response
of the nominal system.
At this stage, it is worthwhile to mention that 
a more realistic infinite-dimensional system expressed by using
partial differential equations
can be discretized into finite-dimensional systems by using
standard numerical techniques such as the Galerkin method.
These discretized systems are often represented by SDOF systems (as in \autoref{eq:eom})
using orthogonal transformations.

The nominal system discussed in \autoref{eq:eom}  has fixed system paramaters  $m_0$, $c_0$ and $k_0$. For a digital twin, however,  the system parameters, namely mass, damping and stiffness, and the forcing function changes with the service time $t_s$. A generalized equation of motion of this system can be represented as
\begin{equation}\label{eq:eom_dt}
    m(t_s) \partialsecondderi{u(t,t_s)}{t} + c(t_s) \partialfirstderi{u(t,t_s)}{t} + k(t_s) u(t,t_s) = f(t,t_s).
\end{equation}
It is to be noted that the service time $t_s$ is much slower than the intrinsic time $t$.
The nominal system discussed in \autoref{eq:eom} can be viewed as the initial model 
at $t_s = 0$. The service time $t_s$ can represent the number of cycles in a aircraft.
From \autoref{eq:eom_dt}, we note that the mass $m(t_s)$, damping $c(t_s)$, stiffness $k(t_s)$ and $f(t,t_s)$ changes with the `service time' $t_s$, for instance due to the degradation 
in the system during its service time.
\autoref{eq:eom_dt} represents the the equation of motion of the digital twin.
Note that when $t_s = 0$, \autoref{eq:eom_dt} reduces to the nominal system represented in \autoref{eq:eom}.
It is evident that the digital twin is completely described by 
the functions $m(t_s)$, $c(t_s)$ and $k(t_s)$.
Therefore, for using the digital twin in practice, one needs to
estimate the functions $m(t_s)$, $c(t_s)$ and $k(t_s)$.

In recent studies, physics-based \cite{Ganguli2020} and data-based 
approaches \cite{chakraborty2020role} for estimating the functions $m(t_s)$, $c(t_s)$ and $k(t_s)$ have been developed.
However, these studies have a number of limitations.
\begin{itemize}
    \item The physics-based digital twin proposed in \cite{Ganguli2020} is not sufficiently accurate when the sensor data is noisy.
    \item The data-based digital twin proposed in \cite{chakraborty2020role} only works for systems having a single operational time-scale. The approach is not applicable for multi-timescale dynamical systems \cite{koutsourelakis2011scalable,Chakraborty2018efficient}.
    \item One of the objectives of digital twin is to predict the future response, so as to understand the behaviour of the physical twin in future. Unfortunately, neither the physics-based \cite{Ganguli2020} nor the data based \cite{chakraborty2020role} digital twins previously proposed is capable of predicting the future responses. 
\end{itemize}
The objective of this study is to develop an efficient framework for addressing some of the above-mentioned limitations.
More specifically, we are interested in developing digital twins for
multi-timescale dynamical systems. 
Unlike the digital twins developed in \cite{Ganguli2020,chakraborty2020role}, the digital twin developed in this paper should
also, be able to predict future responses.

\begin{figure}[ht!]
    \centering
    \subfigure[Mass degradation]{
    \includegraphics[width = 0.6\textwidth]{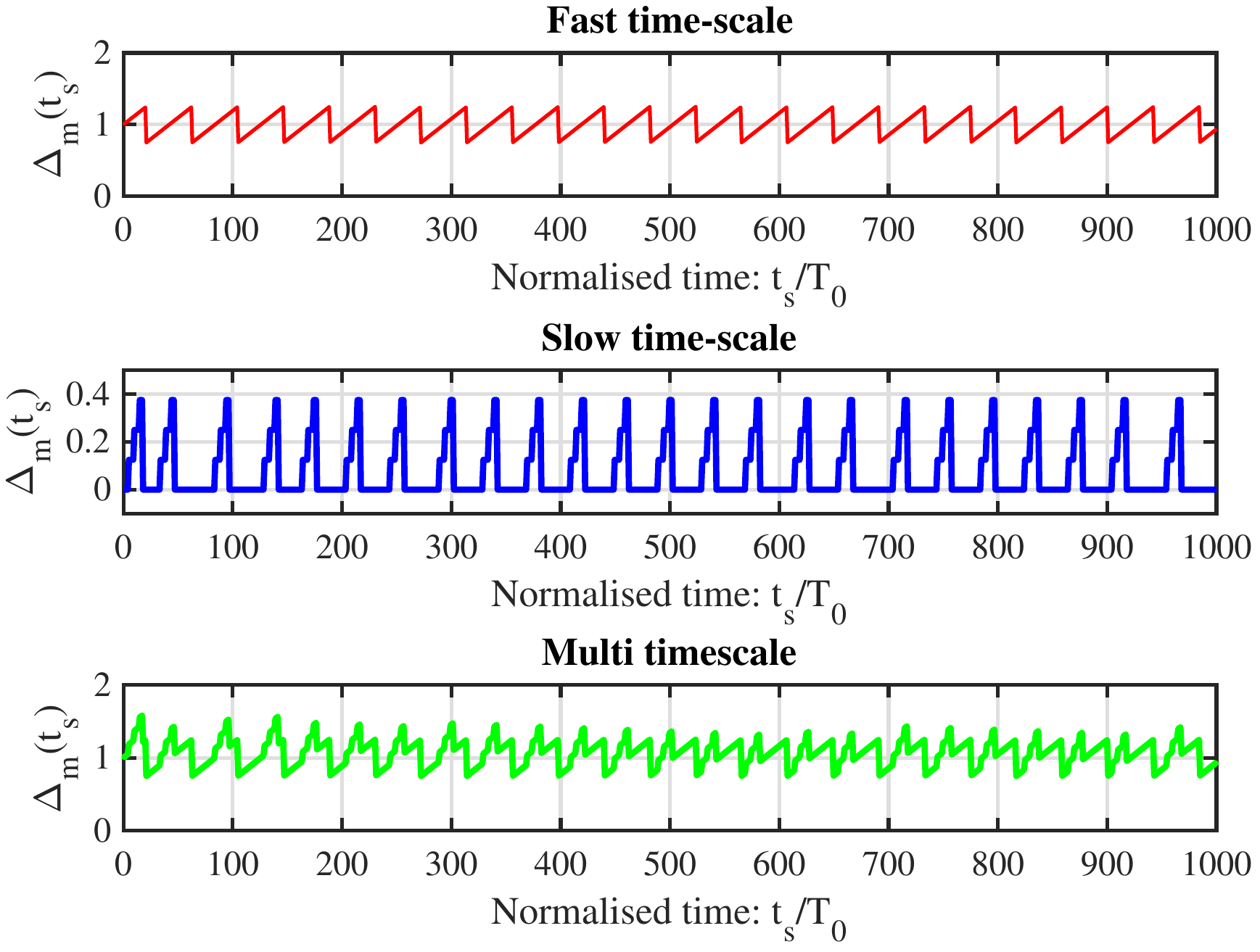}}
    \subfigure[Stiffness degradation]{
    \includegraphics[width = 0.6\textwidth]{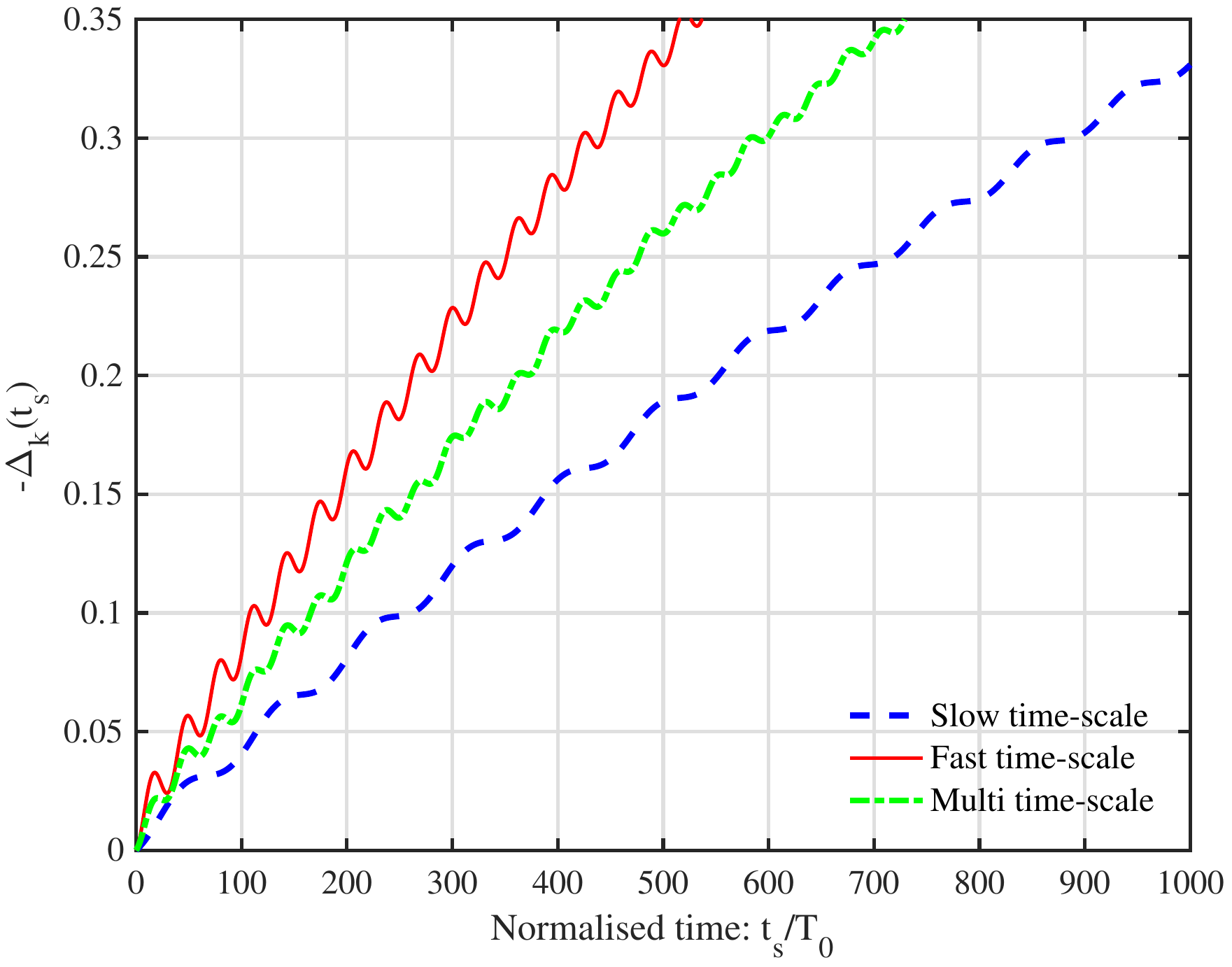}}
    \caption{Multi-scale mass and stiffness degradation functions. The multi-scale degradation functions are obtained by combining the fast and the slow time-scales shown in each figure.}
    \label{fig:degrad}
\end{figure}

For developing the digital twin, it is assumed that sensors are 
deployed on the physical system. 
Recent developments in the field of Internet of Things (IoT) 
has provided us with numerous new data
collection technologies and this
provides the necessary connectivity between the physical
and digital twins.
Using the sensors, measurements are taken intermittently at $t_s$.
It is assumed that the functions $m(t_s)$, $c(t_s)$ and $k(t_s)$ are so slow that the dynamics of the system in \autoref{eq:eom_dt} is decoupled.
In other words, $m_s$, $c_s$ and $k_s$ of the system is constant
as far as the instantaneous dynamics of the system is concerned.
Without loss of generality, we assume

\begin{equation}\label{eq:stiff_vary1}
    k_s(t_s) = k_s\left(t^{(s)},t^{(f)} \right) = k_0 \left(1 + \Delta_k \left(t^{(s)},t^{(f)} \right)\right)
\end{equation}
where
\begin{equation}\label{eq:stiff_degrad1}
 \Delta_k \left(t ^{(s)},t ^{(f)} \right) = \Delta_k^{(s)} \left( t ^{(s)}\right) + \Delta_k ^{(f)}\left(t ^{(f)}\right)  - 1
\end{equation}
Here $t^{(s)}$ and $t^{(f)}$ represent a slower and a faster time scale of evolution of the respective processes. Without any loss of generality, we express these two different time scales as a function of a  single service time-scale $t_s$ with different coefficients. Using this approach we have
\begin{equation}\label{eq:stiff_degrad}
\begin{split}
    \Delta_k \left(t_s\right) & = \Delta_k^{(s)} \left(t_s\right) + \Delta_k ^{(f}) \left(t_s\right) - 1 \\
    & = \underbrace{0.5 e^{-\alpha_k^{(s)} t_s} {\frac{(1+ \epsilon_k^{(s)} \cos(\beta_k^{(s)} t_s))} {(1+\epsilon_k^{(s)})} }}_{{\Delta _k^{(s)}(t^{(s)})}} + \underbrace{0.5 e^{-\alpha_k^{(f)} t_s} {\frac{(1+ \epsilon_k^{(f)} \cos(\beta_k^{(f)} t_s))} {(1+\epsilon_k^{(f)})} }}_{{\Delta _k^{(f)}(t^{(f)})}} - 1. 
\end{split}
\end{equation}
In \autoref{eq:stiff_degrad}, we have assumed that the 
stiffness degradation results from two different processes - one relatively slow and one relatively fast.
Numerical values considered for stiffness degradation are: $\alpha_k^{(s)} = 0.4 \times 10^{-3}$, $\epsilon_k^{(s)} = 0.005$, $\beta_k^{(s)} = 7 \times 10^{-2}$,
$\alpha_k^{(f)} = 0.8 \times 10^{-3}$, $\epsilon_k^{(f)} = 0.01$ and $\beta_k^{(f)} = 2 \times 10^{-1}$.
Similarly, we also assume
\begin{equation}\label{eq:mass_vary}
    m\left(t_s\right) = m_0 \left(1 + \Delta_m \left(t_s\right)\right),
\end{equation}
where
\begin{equation}\label{eq:mass_degrad}
    \Delta_m \left(t_s\right) = \Delta_m^{(s)} \left(t_s\right) + \Delta_m ^{(f)} \left(t_s\right).
\end{equation}
Similar to the stiffness degradation case, the mass degradation is also a function of two time-scales - the relatively slower time-scale $\Delta_m^{(s)} \left(t_s\right)$
and the relatively faster time-scale $\Delta_m^{(f)} \left(t_s\right)$.
We have assumed,
\begin{equation}\label{eq:mass_degrad_fast}
    \Delta_m^{(f)} \left(t_s\right) = \epsilon_m \text{ SawTooth} (\beta_m ( t_s - \pi/\beta_m) ),
\end{equation}
where $\beta_m=0.15$ and  $\epsilon_m=0.25$. 
The slower time-scale is represented as
\begin{equation}\label{eq:mass_degrad_slow}
    \Delta_m^{(s)} \left(t_s\right) = \left\{ \begin{array}{ll}
         1 & \text{if } t_1 \le t_s < t_2 \\ 
         2 & \text{if } t_2 \le t_s < t_3 \\
         3 & \text{if } t_3 \le t_s < t_4 \\
         0 & \text{elsewhere}
    \end{array} \right. .
\end{equation}
From a physical point-of-view, \autoref{eq:mass_degrad_fast}
can be associated with fuel loading and unloading of an aircraft.
On the other hand,
\autoref{eq:mass_degrad_slow} can be associated with the case where the aircraft drops a bomb during its flight.
Schematically, the mass and stiffness degradation are shown in Fig. \ref{fig:degrad}.
The damping is considered to be constant.
The key consideration
is that a digital twin of the dynamical system should be able to track these kinds of
changes occurring at multiple-scales by exploiting sensor data measured on the system.
Moreover, a digital twin should also be able to predict future degradation.

\section{Digital twin for multi-timescale dynamical systems}
\label{sec:dt_ms}

\begin{figure}[ht!]
    \centering
    \includegraphics[width = 1.0\textwidth]{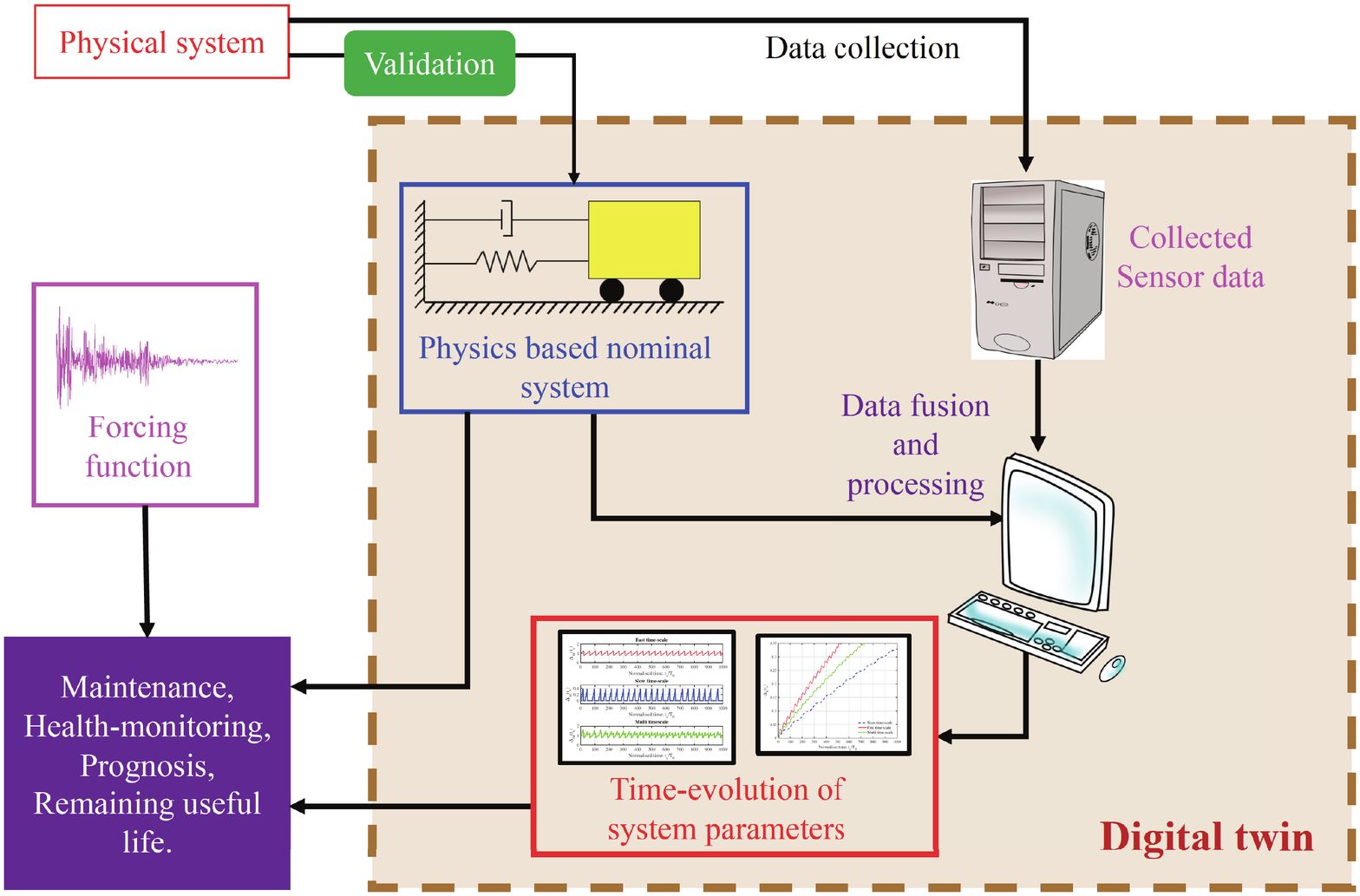}
    \caption{Schematic representation of the digital twin. It has three primary building blocks, namely data fusion and processing, determining time evolution of the system parameters and making predictions using the digital twin. This digital twin can be used for several tasks including prognosis, health-monitoring, maintenance and remaining useful life prediction.}
    \label{fig:dt}
\end{figure}

In this section, we discuss the proposed digital twin framework
for multi-timescale dynamical systems.
A schematic representation of the framework is shown in Fig. \ref{fig:dt}.
The framework proposed has two primary components - (a) data processing by using the physics of the problem (physics-based nominal model) and 
(b) Learning the time-evolution of system parameters by using 
machine learning (ML).
Once the material degradation is known, the future responses
can be predicted by combining the ML predicted material properties
with the physics of the problem defined by the governing differential equation.
To track the multi-scale nature of the degradation functions, we propose to use the concept of mixture of experts
(MOE) where each expert is employed to track a single time-scale.
Based on the success of the Gaussian process (GP) in solving problems
having single time-scale \cite{chakraborty2020role}, we propose to use GP as the experts within the MOE framework.
The overall framework is referred to as the mixture of experts using Gaussian process (ME-GP).
We first present the details on data processing and then proceed to discuss the concept of the proposed ME-GP.

\subsection{Data collection and processing}
\label{subsec:data}
One major player in the development of the digital twin technology is the IoT.
Advances in IoT have provided us with several new data collection technologies that, in turn, drives the
development of the digital twin technology and enables
connectivity between and physical and the digital twins.
The overall idea of the digital twin technology is based on 
the idea of this connectivity.
This connectivity is established by placing sensors on the physical twin to collect data and communicate it to the 
digital counterpart by using cloud technology.
With advances in the sensor technologies, we now have different 
sensors for collecting different type of responses.
In this work, we work with the natural frequency of the system.
The advantage of using natural frequency resides in the fact that it is a scalar quantity and hence, we can avoid working with a big data-set.
We assume that the frequency of the system can be measured in an
online fashion.
Available literature illustrates that this is feasible.
In \cite{feng15}, a vision-based sensor capable of remotely measuring the structural response was proposed.
The effectiveness of the proposed sensor was illustrated by conducting field tests on railway bridges in both time and
frequency domains.
A sensor that infers the natural frequency of a system from vibration
induced strain was proposed in \cite{wang2016dynamic}.
Applicability of this sensor was illustrated by conducting 
experiments on metal pipe under vibration and impact load.
Electrical strain gauge, piezoelectric accelerometer and 
fibre Bragg gratting sensors were used for obtaining the 
natural frequency of the system.
Both these studies show that measuring the natural frequency of a system is feasible, a fact that we use in our study.

In this paper, three different cases have been considered.
In the first case, it is assumed that
only the stiffness degrades.
In the second case, we assume the stiffness to be constant;
the variability in the observations is due to the variation
in the mass.
Lastly, in the third case, we assume that both mass and stiffness vary.
The collected data needs to be processed differently for
each of these three cases.
Details on how the data is processed for each of these three cases are furnished below.
\subsubsection{Stiffness degradation}
\label{subsubsec:stiff_degrad}
We assume that the mass and damping of the nominal model in \autoref{eq:eom_dt} are unchanged and only the 
stiffness degrades. Accordingly,
the equation of motion for this case is written as
\begin{equation}\label{eq:eom_dt_stff_degrad}
    m_0 \frac{\text{d}^2 u(t)}{\text{d} t^2} + c_0 \frac{\text{d} u(t)}{\text{d} t} + k_0\left( 1 + \Delta_k (t_s)\right) u(t) = f(t).
\end{equation}
where all the terms have similar notations as defined before.
Note that \autoref{eq:eom_dt_stff_degrad} is a special case of \autoref{eq:eom_dt} where $t_s$ is fixed.
Solving the characteristic equation, the damped 
natural frequency of the system can be represented as
\begin{equation}\label{eq:lambda_stff_only}
    \lambda_{s_{1,2}} (t_s) = - \zeta_0 \omega_0 \pm \text{i} \omega_0 \sqrt{1 + \Delta_k(t_s) = \zeta_0^2},
\end{equation}
where $\omega_0$ and $\zeta_0$ are respectively the natural
frequency and damping ratio of the system at $t_s = 0$.
\autoref{eq:lambda_stff_only} can be rearranged as
\begin{equation}
    \lambda_{s_{1,2}} \left(t_s\right) = - \underbrace{\frac{\zeta_0}{\sqrt{1 + \Delta_{\hat k}\left(t_s\right)}}}_{\zeta_s\left(t_s\right)} \underbrace{\omega_0 \sqrt{1 + \Delta_{\hat k}\left(t_s\right)}}_{\omega_s\left(t_s\right)} \pm \text{i} \underbrace{\omega_0\sqrt{1 + \Delta_{\hat k}\left(t_s\right)}\sqrt{1 - \left(\frac{\zeta_0}{\sqrt{1 + \Delta_{\hat k} \left( t_s \right)}}\right)^2}}_{\omega_{d_s}\left(t_s\right)},
\end{equation}
where 
$\omega_s\left(t_s\right) = \omega_0 \sqrt{1 + \Delta_{\hat k}\left(t_s\right)}$ i, 
$\zeta\left(t_s\right) = \zeta_0 /\sqrt{1 + \Delta_{\hat k}\left(t_s\right)}$ 
and 
$\omega_{d_s}\left(t_s\right) = \omega_s \left(t_s\right) \sqrt{1 - \zeta_s^2 \left( t_s \right)}$ 
represent the evolution of the natural frequency, damping ratio
and damped natural frequency with $t_s$.
As the natural frequency extraction techniques in literature generally extract
the damped natural frequency, we have considered it to be the data
available from the physical twin.
It can be shown \cite{Ganguli2020}
\begin{equation}\label{eq:delta_k}
    \Delta_{\hat k}\left(t_s \right) = - \tilde d_1 \left(t_s\right)\left(2 \sqrt{1 - \zeta_0^2} - \tilde d_1\left(t_s \right) \right),
\end{equation}
where
\begin{equation}\label{eq:abs_dist}
    \tilde d_1 \left(t_s \right) = \frac{d_1\left(\omega_{d_0}, \omega_{d_s}\left(t_s\right) \right)}{\omega_0}.
\end{equation}
The function $d_1\left( \omega_{d_0}, \omega_{d_s}\left(t_s\right) \right)$ in \autoref{eq:abs_dist} is the 
distance between $\omega_{d_0}$ and $\omega_{d_s}\left(t_s\right)$
\begin{equation}
    d_1\left( \omega_{d_0}, \omega_{d_s}\left(t_s\right) \right) = \left||\omega_{d_0} -  \omega_{d_s}\left(t_s\right) \right||_2.
\end{equation}
Now given the fact that the initial damped frequency of the system, $\omega_{d_0}$ is known and we have sensor measurements for $\omega_{d_s}\left(t_s\right)$, one can easily compute $\tilde d_1 \left(t_s \right)$ and $\Delta_{\hat k} \left(t_s\right)$ at $t_s$ by using \autoref{eq:abs_dist} and substituting it into \autoref{eq:delta_k}.
Note that the sensor measurements $\tilde d_1 \left(t_s \right)$ are likely to be corrupted by noise and hence, the estimates for $\Delta_{\hat k}$ are also noisy. 
In this study, this noisy estimates, $\Delta_{\hat k}$ at discrete time $t_s$ are used for 
developing the digital twin for the multi-timescale dynamical system.
\subsubsection{Mass evolution}
\label{subsubsec:mass_degrad}
In this case, we consider that the stiffness and damping of the 
nominal model in \autoref{eq:eom_dt} are constant, and the variation in the observed natural frequency is 
due to variation in the mass during the service life.
Accordingly, the equation of motion of the physical system reduces to
\begin{equation}\label{eq:eom_dt_mass_degrad}
    m_0\left( 1 + \Delta_m (t_s) \right) \frac{\text d^2 u(t)}{\text d t^2} + c_0 \frac{\text d u(t)}{\text d t} + k_0 u(t) = f (t).
\end{equation}
Again, \autoref{eq:eom_dt_mass_degrad} is a special case \autoref{eq:eom_dt} where only $m$ varies and the stiffness is constant.
Solving for the damped natural eigenfrequencies as before
\begin{equation}\label{eq:nat_freq_mass}
    \lambda_{s_{1,2}} \left(t_s\right) = - \omega_s\left(t_s\right)\zeta_s\left(t_s\right) \pm \text i \omega_{d_s}\left(t_s\right),
\end{equation}
where
\begin{subequations}
    \begin{equation}
        \omega_s\left(t_s\right) = \frac{\omega_0}{\sqrt{1 + \Delta_{\hat m}\left(t_s\right)}},
    \end{equation}
    \begin{equation}
        \zeta_s\left(t_s\right) = \frac{\zeta_0}{\sqrt{1 + \Delta_{\hat m}\left(t_s\right)}}\;\; \text{and}
    \end{equation}
    \begin{equation}
        \omega_{d_s}\left(t_s\right) = \omega_s \left(t_s\right)\sqrt{1 - \zeta_s^2\left(t_s\right)}.
    \end{equation}
\end{subequations}
are the evolution of natural frequency, damping ratio and damped natural frequency of the digital twin.
Similar to the stiffness degradation case, we have
\begin{equation}\label{eq:mass_cf}
\begin{split}
    \Delta_{\hat m}\left(t_s\right) = & \frac{-2\tilde d_2\left(t_s\right)^2 + 4\tilde d_2\left(t_s\right)\sqrt{1 - \zeta_0^2} - 1 + 2\zeta_0^2 }{2 \left( - \tilde d_2 \left(t_s\right) + \sqrt{1 - \zeta_0^2} \right)^2} \\
   & + \frac{\sqrt{1 - 4\tilde d_2\left(t_s\right)^2 \zeta_0^2 + 8 \tilde d_2\left(t_s\right) \sqrt{1 - \zeta_0^2}\zeta_0^2 - 4\zeta_0^2 + 4 \zeta_0^4}}{2 \left( - \tilde d_2 \left(t_s\right) + \sqrt{1 - \zeta_0^2} \right)^2}
\end{split}
\end{equation}
where $\tilde d_2 \left(t_s\right)$ is the equivalent of $\tilde d_1$ for the stiffness evolution case. 
Again we emphasize that the sensor based estimates of the damped natural frequencies are noisy and hence, the estimated 
$\Delta_{\hat m}\left(t_s\right)$ are also noisy.
In this case, we utilize the noisy $\Delta_{\hat m}\left(t_s\right)$ at discrete time $t_s$ for developing the
digital twin for multi-scale systems.
\subsubsection{Mass and stiffness evolution}
\label{subsubsec:mass_and_stiff_degrad}
In this case, we consider the evolution of mass and degradation
of stiffness, simultaneously.
The equation of motion in this case is represented as
\begin{equation}\label{eq:eom_dt_mas_stiffness_degrad}
    m_0\left(1 + \Delta_m (t_s)\right) \frac{\text d ^2 u(t)}{\text d t^2} + c_0 \frac{\text d u(t)}{\text d t} + k_0 \left( 1 + \Delta_k (t_s) \right) u(t) = f(t).
\end{equation}
All the notations in \autoref{eq:eom_dt_mas_stiffness_degrad}
have same meaning as before.
The damped natural eigenfrequencies of this system can are represented as
\begin{equation}\label{eq:freq}
    \lambda_{s_{1,2}} = -\omega_s \left(t_s\right) \zeta_s\left(t_s\right) \pm \text{i} \omega_{d_s}\left(t_s\right),
\end{equation}
where
\begin{subequations}\label{eq:dynamic_qoi}
    \begin{equation}
        \omega_s \left(t_s\right) = \omega_0 \frac{\sqrt{1 + \Delta_{\hat k}\left(t_s\right)}}{\sqrt{1 + \Delta_{\hat m}\left(t_s\right)}}
    \end{equation}
    \begin{equation}
        \zeta_s\left(t_s\right) = \frac{\zeta_0}{\sqrt{1 + \Delta_{\hat m}\left(t_s\right)}\sqrt{1 + \Delta_{\hat k}\left(t_s\right)}}\;\;\text{and}
    \end{equation}
    \begin{equation}
        \omega_{d_s} \left(t_s\right) = \omega_{s} \left(t_s\right)\sqrt{1 - \zeta_s^2}.
    \end{equation}
\end{subequations}
$\omega_s$, $\zeta_s$ and $\omega_{d_s}$, respectively represent
the evolution of natural frequency, damping ratio and 
damped natural frequency.
Unlike the previous two cases, both $\Delta_m(t_s)$ and $\Delta_k(t_s)$ are unknowns in this case and hence,
we need two equations to solve these unknowns.
To that end, we consider the real and imaginary parts of 
\autoref{eq:freq} separately to derive the two equations
necessary for estimating $\Delta_m(t_s)$ and $\Delta_k(t_s)$.
With this setup, we arrive at the following expression \cite{Ganguli2020}
\begin{subequations}\label{eq:delta_mk}
    \begin{equation}
        \Delta_{\hat m} \left(t_s\right) = - \frac{\tilde d_{\mathcal R} \left(t_s\right)}{\zeta_0 + \tilde d_{\mathcal R} \left(t_s\right)},
    \end{equation}
    \begin{equation}
        \Delta_{\hat k} \left(t_s\right) = \frac{\zeta_o \tilde d_{\mathcal R}^2 \left(t_s\right) - \left( 1 - 2 \zeta_0^2 \right) \tilde d_{\mathcal I} \left(t_s\right) + \zeta_0^2 \tilde d_{\mathcal I}^2 \left(t_s\right)}{\zeta_0 + \tilde d_{\mathcal R}\left(t_s\right)},
    \end{equation}
\end{subequations}
where $\tilde d_{\mathcal R}\left(t_s\right)$ and $\tilde d_{\mathcal I}\left(t_s\right)$, as before, are distance measures
\begin{equation}
    \tilde d_{\mathcal R}\left(t_s\right) = \frac{d_{\mathcal R}\left(t_s\right)}{1 + \Delta_{\hat m}\left(t_s\right)},\;\;\;\;\; \tilde d_{\mathcal I}\left(t_s\right) = \sqrt{1 - \zeta_0^2} - \frac{\sqrt{\left( 1 + \Delta _{\hat k}\left(t_s\right) \right)\left( 1 + \Delta _{\hat m}\left(t_s\right) \right) - \zeta_0^2}}{1 + \Delta_{\hat m}\left(t_s\right)}.
\end{equation}
Note that $\Delta_{\hat m} \left(t_s\right)$ and $\Delta_{\hat k} \left(t_s\right)$
are estimated from noisy observations of $\lambda$ and hence, are noisy.
In this case also, we will utilize $\Delta_{\hat m} \left(t_s\right) $ and $\Delta_{\hat k} \left(t_s\right) $
obtained at discrete time $t_s$ within the digital twin framework for multi-timescale dynamical system.
\subsection{Mixture of experts with Gaussian process}
\label{subsec:me-gp}
In the next part of the digital twin framework, we use the 
processed data to learn the evolution of the system parameters.
Note that the evolution of the system parameters are of 
multi-scale nature and hence, is difficult to learn.
In this paper, we propose the use of mixture of experts (MoE) within the digital twin framework.  
MoE is used to learn the evolution of the system parameters.
We argue that each of the experts within MoE learns the 
evolution at a single scale and hence, MoE can predict 
the multi-scale evolution of the system parameters.
As experts within the MoE framework, we propose the use of GP.
The effectiveness of GP in predicting parameter evolution at
a single scale has already been established in a previous study \cite{chakraborty2020role}.

Suppose, we have a sequence of observations $\bm y_{t_s} \in \mathbb R^d$ at discrete time $t_s$, $s = 1,2,\ldots,\tau$.
For the digital twin problem in this study, $\bm y_{t_s}$ can either be $\Delta_k (t_s)$ and/or $\Delta m (t_s)$ at discrete time $t_s$.
This observations are generated from some unknown process
having multiple time-scales.
We assume that the observations are generated from $M$  hidden states $x_t^{(m)}$, $m=1,2,\ldots,M$, also referred to as experts.
This hidden states are generally assumed to be independent and can evolve independent of each other.
We here assume that the independent hidden states evolves
according to a GP
\begin{equation}\label{eq:hidden_state_GP}
    x_m | t \sim \mathcal{GP}\left( \mu _m (t; \bm h), \kappa _m (t_1, t_2; \bm {l}\right), \;\;\;\; m = 1,2,\ldots, M,
\end{equation}
where
\begin{equation}\label{eq:mean_func_GP}
    \mu _m (t) = \bm h ^T \bm \phi (t), 
\end{equation}
represents the mean of the GP. 
$\bm h$ in \autoref{eq:hidden_state_GP} represents the unknown coefficients and $\bm \phi (t)$ represents the basis function vector.
$\kappa _m$ in \autoref{eq:hidden_state_GP} represents the correlation function with length-scale parameter $\bm l$.
$\bm h$ and $\bm {l}$ are referred to as the hyperparameters of the GP.
\autoref{eq:hidden_state_GP} can be viewed as the prior, parameterized by the hyperparameters $\bm h$ and $\bm {l}$ in the space of the hidden state.
These hidden states can be coupled in a generative manner
to obtain the underlying data
\begin{equation}\label{eq:output_ME_GP}
    y_{t_s} = y(t = t_s) = \sum_{m=1}^M z_{m}\left(t;\bm \theta^g\right) x_m \left(t;\bm \theta^e\right).
\end{equation}
$z_m \left( t \right)$ is the $m$--th gating function and defined as
\begin{equation}\label{eq:gating}
    z_i(t) = \frac{\pi_i \mathcal N \left(t|\mu_i, \lambda_i^{-1} \right)}{\sum_{j=1}^M{\pi_j \mathcal N \left(t|\mu_j, \lambda_j^{-1} \right)}},\;\;\;\;\sum_{j=1}^M \pi_j = 1.
\end{equation}
$\bm \theta^g = \left[ \mu_j, \lambda_j \right]_{j=1}^M$ are the hyperparameters of the gating function.
$\pi_j$, $j=1,2,\ldots,M$ in \autoref{eq:gating} represents the mixing coefficient.
$x_m$ in  \autoref{eq:output_ME_GP}  is the  $m$-th GP expert. $\bm \theta^e = \left[ \bm h_m, \bm {l}_m \right]_{m=1}^M$ is the hyperparameter associated with the expert function.

For using the model defined in Eqs.  (\ref{eq:hidden_state_GP})--(\ref{eq:gating}),
all the hyperparameters need to be estimated based on the training data $\mathcal D = \left[\bm y_{t_s}, t_s \right]$.
One way to achieve this is by maximizing the data likelihood of the model.
\begin{equation}\label{eq:data_likelihood}
    p\left(y_{t_s} | t_s, \bm \theta, \bm \pi \right) = \sum_{i=1}^M {p\left(i| t_s, \bm \theta ^g, \bm \pi\right)p\left( y_{t_s}| t_s,\bm \theta ^e, \right)}
\end{equation}
$p\left(i| t_s, \bm \theta ^g, \bm \pi\right)$ in
\autoref{eq:data_likelihood} is the posterior conditional probability, where $ t_s $ is assigned to the 
partition corresponding to the $i$--th expert, i.e., 
\begin{equation}
    p\left(i| t_s, \bm \theta ^g, \bm \pi\right) = z_i(t).
\end{equation}
$p\left( \bm y_{t_s}| t_s,\bm \theta ^e \right)$, on the other hand, is the probability distribution of the $i$--th expert and hence is a GP
\begin{equation}\label{eq:likelihood_expert}
    p\left(y_{t_s}| t_s,\bm \theta ^e, \right)  = \mathcal N \left( \mu_i \left( t_s;\bm h \right), \kappa _i \left( t_{s,1},  t_{s,2}; \bm l  \right) \right).
\end{equation}
Substituting Eqs. (\ref{eq:gating}) and (\ref{eq:likelihood_expert}) into \autoref{eq:data_likelihood}, we obtain
\begin{equation}\label{eq:data_likelihood2}
    p\left(y_{t_s} | t_s, \bm \theta, \bm \pi \right) = \sum_{i=1}^M {\frac{\pi_i \mathcal N \left(t|\mu_i, \lambda_i^{-1} \right)}{\sum_{j=1}^M{\pi_j \mathcal N \left(t|\mu_j, \lambda_j^{-1} \right)}}\mathcal N \left( y_{t_s} | \mu_i \left( t_s;\bm h \right), \kappa _i \left( t_{s,1},  t_{s,2}; \bm l  \right) \right)}.
\end{equation}
Note that \autoref{eq:data_likelihood2} is analytically
intractable. 
Using the training samples $\bm y_{t_s}$ and $\bm t_s$, $s = 1,2,\ldots, \tau$, the likelihood can be represented as
\begin{equation}\label{eq:data_likelihood3}
    p\left(\bm y_{t_s} | \bm t_s, \bm \theta, \bm \pi \right) = \prod_{s=1}^{\tau}\sum_{i=1}^M {\frac{\pi_i \mathcal N \left(t_s|\mu_i, \lambda_i^{-1} \right)}{\sum_{j=1}^M{\pi_j \mathcal N \left(t_s|\mu_j, \lambda_j^{-1} \right)}}\mathcal N \left(y_{t,s} | \mu_i \left( t_s;\bm h \right), \kappa _i \left( t_{s,1},  t_{s,2}; \bm l  \right) \right)}.
\end{equation}
One way to estimate the parameters in \autoref{eq:data_likelihood3} is by using maximum likelihood estimator where we maximize the 
likelihood in \autoref{eq:data_likelihood3}.
However, such an approach often leads to over-fitting.
An alternative to the maximum likelihood estimator is to
using an Bayesian approach and compute the posterior distribution of the hyperparameters.
However, as the likelihood is intractable for the problem
in hand, such an approach is computationally expensive.
In this paper, with adopt a hybrid approach where some of the parameters are treated in a Bayesian way while for the other parameters, point estimates are computed.
More specifically, within the proposed framework, we compute
point-estimates for the mixing coefficients $\bm \pi$.
The hyperparameters corresponding to the gating distribution and the experts are treated in a Bayesian way.

To estimate the hyperparameters using the proposed hybrid
approach, we first use Bayes rule to compute the 
posterior distribution of the hyperparameters $\bm \theta$ and $\bm \pi$
\begin{equation}\label{eq:posterior_hp}
    p\left(\bm \theta, \bm \pi | \bm y_s, \bm t_s \right) = \frac{p\left( \bm \pi, \bm \theta \right) p \left(\bm y_s | \bm t_s, \bm \pi, \bm \pi \right)}{p\left(\bm y_s | \bm t_s \right)},
\end{equation}
where $p\left( \bm \pi, \bm \theta \right) $ represents prior distribution of the hyperparameters and
$p \left(\bm y_s | \bm t_s, \bm \pi, \bm \pi \right)$ is obtained from \autoref{eq:data_likelihood3}.
Recall that the goal is to compute point-estimates for the mixing parameters $\bm \pi$.
This can be achieved by maximizing the log-posterior for the mixing coefficients
\begin{equation}\label{eq:log_posterior}
    \mathcal L\left(\bm \pi \right) = \log p\left( \bm \pi | \bm y_{t_s}, \bm t_s \right) = \log \int p\left( \bm \pi , \bm \theta | \bm y_{t_s}, \bm t_s \right)\text d \bm \theta.
\end{equation}
Unfortunately, this is not straightforward as \autoref{eq:log_posterior} involves integration over the
unknown $\bm \theta$.
In this work, we propose to use expectation maximization for computing the mixing coefficients by maximizing the
log-posterior in \autoref{eq:log_posterior}.
In expectation maximization, we iterate over a series of increasing lower-bound of $\mathcal L\left(\bm \pi \right)$ by using the Jensen's inequality
\begin{equation}\label{eq:jensen}
\begin{split}
    \mathcal L\left(\bm \pi \right) = ) = \log p\left( \bm \pi | \bm y_{t_s}, \bm t_s \right) & = \log \int p\left( \bm \pi , \bm \theta | \bm y_{t_s}, \bm t_s \right)\text d \bm \theta \\
    & = \log \int q\left( \bm \theta \right)
    \frac{p\left( \bm \pi , \bm \theta | \bm y_{t_s}, \bm t_s  \right)}{q\left( \bm \theta \right)}\text d \bm \theta \\
    & \ge \int q\left( \bm \theta \right) \log \frac{p\left( \bm \pi , \bm \theta | \bm y_{t_s}, \bm t_s  \right)}{q\left( \bm \theta \right)}\text d \bm \theta \\
    & = F \left(q,\bm\pi\right),
\end{split}
\end{equation}
where $q\left( \bm \theta \right)$ is an auxiliary distribution. It is obvious that the equality in \autoref{eq:jensen} holds when $q \left( \bm \theta \right) = p\left( \bm \theta | \bm \pi, \bm y_{t_s}, \bm t_s \right)$.
Using expectation maximization, $\bm \pi$ is estimated by iterating over the E-step (expectation step) and the M-step (maximization step).
\begin{itemize}
    \item \textbf{E-step}: Given an estimate of $\bm \pi = \bm \pi^{(s)}$ in step s, we compute the lower-bound
    \begin{equation}\label{eq:E-step}
    \begin{split}
        F\left( q^{(s)}, \bm \pi \right) = & \int q^{(s)} \left( \bm \theta \right) \log p \left( \bm \pi. \bm \theta | \bm y_{t_s}, \bm t_s \right) \text d \bm \theta \\
        & - \int q^{(s)} \left( \bm \theta \right) \log \int q^{(s)} \left( \bm \theta \right) \text d \bm \theta. 
    \end{split}
    \end{equation}
    \item \textbf{M-step}: Maximize $F\left( q^{(s)}, \bm \pi \right)$ to update $\bm \pi$.
    \begin{equation}\label{eq:M-step}
    \begin{split}
        \bm \theta ^{(s+1)} & = \arg \max_{\bm \theta} F\left( q^{(s)}, \bm \pi \right) \\
        & = \arg \max_{\bm \theta} \left[ \mathbb E_{q^{(s)}\left( \bm \theta \right)}\left( \log p\left( \bm \pi, \bm \theta | \bm y_{t_s}, \bm t_s \right) \right) \right].
    \end{split}
    \end{equation}
\end{itemize}
The second equality in \autoref{eq:M-step} holds
because the second term of $F\left( q^{(s)}, \bm \pi 
\right)$ is independent of $\bm \pi$.
It is important to note that the optimal distribution 
$q^{(s)} \left( \bm \theta \right) = p \left( \bm \theta | \pi^{(s)}, \bm y_{t_s}, \bm t_s  \right)$ is intractable.
We propose to use sequential Monte Carlo (SMC) sampler \cite{del2006sequential} to generate samples from $p \left( \bm \theta | \pi^{(s)}, \bm y_{t_s}, \bm t_s  \right)$ so that the expectation in
the E-step can be represented as
\begin{equation}\label{E-step}
    \mathbb E_{q^{(s)}\left( \bm \theta \right)}\left( \log p\left( \bm \pi, \bm \theta | \bm y_{t_s}, \bm t_s \right) \right) \approx \sum_{i=1}^{N_s} W^{(s,i)} \log p\left( \bm \pi^{(s)}, \bm \theta^{(s,i)} | \bm y_{t_s}, \bm t_s \right),
\end{equation}
where $\bm \theta^{(s,i)}$ is the $i$--th sample generated from $p \left( \bm \theta | \pi^{(s)}, \bm y_{t_s}, \bm t_s  \right)$, and $W^{(s,i)}$ is the corresponding weight.

Often posterior distributions are 
multi-modal and conventional Markov Chain Monte Carlo (MCMC)
\cite{Murphy2012} may get trapped in a local mode.
This results in long mixing time making the process inefficient.
One algorithm that addresses this issue is the SMC sampler \cite{del2006sequential,doucet2000sequential}.
SMC provides a parallelizable 
framework for efficiently drawing 
samples from multi-modal posterior
distributions. 
The idea of annealing is introduced to construct auxiliary distributions.
We traverse from the prior to the 
posterior through this auxiliary distributions; 
this ensures a smooth transition
from the tractable prior to the
intractable posterior.
It can be shown that samples drawn using SMC converges asymptotically
to the target distribution \cite{del2006sequential}.

For using SMC to approximate the E-step of the expectation maximization algorithm,
we first express $p \left( \bm \theta | \pi^{(s)}, \bm y_{t_s}, \bm t_s  \right)$ as
\begin{equation}\label{eq:posterior}
    p \left( \bm \theta | \pi^{(s)}, \bm y_{t_s}, \bm t_s  \right) \propto p\left( \bm \theta \right)p\left(\bm y_{t_s}|\bm t_s. \bm \pi^{(s)}, \bm \theta \right).
\end{equation}
where $p\left( \bm \theta \right)$
is the prior and 
$p\left(\bm y_{t_s}|\bm t_s. \bm \pi^{(s)}, \bm \theta \right)$
is the likelihood of the model defined in \autoref{eq:data_likelihood3}.
In this work, we set the prior as a multivariate Gaussian distribution
with zero mean and identity covariance matrix.
Therefore, the parameters $\bm \theta $ are independent in the prior.
For ease of representation, we write the likelihood in a compress form as $p\left( \mathcal D | \bm \theta \right)$ and the posterior of $\bm \theta$ as $p_n (\bm \theta)$. With these notations,
\autoref{eq:posterior} is represented as
\begin{equation}\label{eq:posterior2}
    p_n\left( \bm \theta \right) \propto p\left(\bm \theta \right) p \left( \mathcal D | \bm \theta \right)
\end{equation}
Based on \autoref{eq:posterior2},
we formulate the following auxiliary distribution in SMC
\begin{equation}
    p_t \left( \bm \theta \right) \propto p\left(\bm \theta \right) p ^{\gamma_t} \left( \mathcal D | \bm \theta   \right),
\end{equation}
where $t=0,1,\ldots,n$ and $0=\gamma_0 < \gamma_1 < \cdots < \gamma_n = 1 $ are the annealing parameters. 
Using the SMC sampler, samples are drawn from such a sequence of probability distribution 
by utilizing importance sampling and re-sampling.
At step $t$, the idea is to generate a sufficient collection of 
$\left\{ \bm \theta_r ^{(i)}, \bm w_r ^ {(i)} \right\},\;\;i=1,\ldots,N_s$ such that
the empirical distribution converges asymptotically to the target distribution $p_r \left( \bm \theta \right)$.
Sampling at $t=0$ is trivial (as we sample the prior).
From $t=1$ onward, we employ importance sampling sequentially to the auxiliary distributions.
A predefined Markov transition kernel is used to that end.
Assuming, at step $t-1$, $N_s$ samples $\left\{\bm \theta _{t-1}^{(i)}\right\}$, $i=1,\ldots, N_s$ are generated according to the 
proposal distribution $\varphi _{t-1}$, a kernel $K_t$ with invariant distribution
$p_t$ is proposed such that
the new samples are marginally distributed as \cite{wan2013probabilistic}
\begin{equation}
    \varphi _t = \int \varphi _{t-1} K_t \left( \bm \theta, \bm \theta' \right) \text d \bm \theta.
\end{equation}
Following \cite{del2006sequential}, we have utilized the Metropolis-Hasting
kernel with invariant distribution
$p_t$ to move the samples based on a random walk proposal
\begin{equation}
    \varphi _t = \mathcal{N}\left( \bm \theta _ {r-1}^{(i)}, \mathbf v^{(i)} \right),
\end{equation}
where $\mathbf v^{(i)}$ is the covariance matrix.
To represent the discrepancy between the proposal distribution $\varphi_t$ and the target distribution $p_t$ at step $t$,  $0 < t \le n$, unnormalized
importance weights $w_t^{(i)}$ are generated.
\begin{equation}\label{eq:unnorm_weights}
    w_t^{(i)} = w_t^{(i-1)}\frac{p_t\left( \bm \theta _{t-1}^{i} \right)}{p_{t-1}\left( \bm \theta _{t-1}^{i} \right)}
\end{equation}
The computed weights are normalized as
\begin{equation}\label{eq:norm_weights}
    W_t^{(i)} = \frac{w_t^{(i)}}{\sum_{j=1}^{N_s}w_t^{(j)}}
\end{equation}

As pointed out in \cite{del2006sequential,liu1995blind}, the SMC sampler degenerates and the variance of the importance weight increases.
In this work, we measure the degeneracy based on the effective sample size (ESS) \cite{wan2013probabilistic}
\begin{equation}\label{eq:ess}
    \text {ESS}_t = \left( \sum_{t=1}^{N_s} \left( W_t^{(i)} \right)^2 \right)^{-1}.
\end{equation}
We consider degeneracy to have occurred if 
\begin{equation}
  \text {ESS}_t < \text {ESS}_{\text{min}},  
\end{equation}
where $\text {ESS}_{\text{min}}$ represents the threshold.
In this work, we have defined $\text {ESS}_{\text{min}} = c \times N_s \left( c < 1 \right)$.
In case, $\text {ESS}_t < \text {ESS}_{\text{min}}$, resampling is carried out to relieve the degeneracy of the sampler.
Once samples corresponding to the target distribution are obtained,
we utilize them to compute the expectation in the E-step of the expectation maximization algorithm.
The steps involved in the SMC sampler are shown in Algorithm \ref{alg:smc}.
\begin{algorithm}[ht!]
\caption{Sequential Monte Carlo sampler}
\label{alg:smc}
\textbf{Input: }Number of samples to generate $N_s$, the prior distribution $p\left( \bm \theta \right)$, the number of steps $n$ and the threshold parameters $c$ \\
Initialize $N_s$ particles $\bm \theta_0^{(i)}$, $i=1,\ldots,N_s$ by directly sampling the prior distribution $p\left(\bm \theta \right)$ and set the corresponds weights to be one, $w_0^{(i)} = 1$.  \\
\For{$t=1,\ldots,n$}{
\For{$i=1,\ldots,N_s$}{
Sample $\bm u_i$ from uniform distribtuion $\mathcal U \left( \bm 0, \mathbf I \right)$.\\
Sample $\tilde {\bm \theta}$ from the proposal distribution $\mathcal N \left( \bm \theta _{t-1}^{(i)}, \mathbf v_i \right)$.\\
\eIf{$\bm u_i < \min \left\{ \frac{p_t \left( \tilde {\bm \theta} \right)}{p_t \left( \bm \theta _{t-1}^{(i)} \right)}\right\}$
}{ $\bm \theta_t^{(i)} \leftarrow \tilde {\bm \theta} $
}
{
$\bm \theta_t^{(i)} \leftarrow \bm \theta_{t-1}^{(i)} $
}
}
Set weights of each particle according to Eqs. (\ref{eq:unnorm_weights}) and (\ref{eq:norm_weights}). \\
Compute $\text{ESS}_t$ using \autoref{eq:ess}. \\
Resample if $\text{ESS}_t < \text{ESS}_{\min}$.
}
Use $\bm \theta_n^{(i)}$ and $W_n^{(i)}$, $i=1,\ldots, N_s$ to compute the expectation in the E-step of the expectation maximization algorithm.
\end{algorithm}
The steps involved in training the proposed mixture of experts using GP algorithm are shown in Algorithm \ref{alg:me-gp}

\begin{algorithm}
\caption{Mixture of experts using Gaussian process}
\label{alg:me-gp}
\textbf{Input:} Number of experts $M$, the training data $\mathcal D = \left[\bm y_{t_s}, \bm t_s \right]$, $s=1,\ldots,\tau$, initial values of mixing coefficients $\bm \pi^{(i)}$ and threshold $\epsilon$.\\
$\bm \pi \leftarrow \bm \pi^{(i)}$. \\
$\lambda = 10\epsilon$. \\
\Repeat{$\lambda  \le \epsilon$
}{
$\bm \pi_s \leftarrow \bm \pi$. \\
Compute $F\left(q^{(s)},\bm \pi \right)$ using SMC sampler (Algorithm \ref{alg:smc}). \\
Update $\bm \pi$  by solving the optimization problem in \autoref{eq:M-step}.\\
Compute error threshold \[ \lambda = \left|| \bm \pi - \bm \pi_s \right||_2 \]
}
\textbf{Outcome:} Optimized $\bm \pi$, $N_s$ samples and corresponding weights from the posterior of $\bm \theta$ $\bm \theta_n^{(i)}$, $W_n^{(i)}$, $i=1,\ldots, N_s$.
\end{algorithm}

Once the hyperparameters of the proposed model are predicted by using the proposed approach, we 
proceed to make predictions using the proposed approach.
Since we use a partially Bayesian approach to obtain the 
hyperparameters $\bm \theta$, it is possible to utilize the same to make {\it probabilistic} predictions.
Suppose, we are interested in obtaining $y_{t^*}$ 
at time-step $t^*$.
This can be obtained by computing the posterior predictive distribution.
\begin{equation}\label{eq:post_pred}
    p\left( y_{t^*} | t^*, \mathcal D, \pi^* \right) = \int_{\bm \theta} \sum_{i=1}^M \underbrace{p\left( i | t^*, \pi_i^*,\theta_i^g \right)}_{\text{gating}}\underbrace{p\left(y^* | t^*, \theta_i^e \right)}_{\text{expert}}\underbrace{p\left( \theta_i^e, \theta_i^g | \mathcal D \right)}_{\text{posterior}} \text d \bm \theta.
\end{equation}
The integral above can be approximated by using Monte Carlo integration.
In particular, we use the samples drawn from the posterior along with the correspond weights and the EM estimate of $\bm \pi^*$ to draw samples from the posterior predictive distribution in \autoref{eq:post_pred}.

\subsection{Algorithm}
\label{subsec:algorithm}
We now proceed to discuss how the components discussed in \autoref{subsec:data} and \autoref{subsec:me-gp}
interacts with each other within the digital twin 
framework shown in \autoref{fig:dt}, and how the digital twin enhanced with ME-GP
can be used for multi-timescale dynamical systems.
Given a physical system, the first step towards developing a digital twin is to develop a 
physics-driven nominal model for the system.
For the current work, the nominal model is represented 
by \autoref{eq:eom_dt}.
Next, the collected responses (damped natural frequencies of the system) are processed by using the procedure discussed in \autoref{subsec:data}.
To be more specific, we process the collected damped natural frequencies to obtain change in the mass, ($\Delta_m(t_s)$) and stiffness ($\Delta_k(t_s)$) of the system.
In the third step, the time-evolution of 
mass and stiffness, $\eta: t \rightarrow \Delta_k, \Delta_m$
is learned by using ME-GP.
Finally, Using the trained ME-GP, we compute the future mass and stiffness, substitute them into the nominal model and solve it to obtain the future responses of interest.
These future responses of interest can be used for
health-monitoring, computing remaining useful life, devising a maintenance strategy and identifying defects and/or cracks in the system.
How the algorithm within digital twin works is shown in Algorithm \ref{alg:dt}.

\begin{algorithm}[ht!]
\caption{Proposed digital twin}
\label{alg:dt}
\textbf{Input:} Nominal model and damped natural frequency of the physical system at different time-instants, $\mathcal D = \left[\lambda_s, t_s \right],\; s = 1,\ldots,\tau$.\\
Process the collected data to obtain $\Delta_k (t_s)$ and/or $\Delta_m (t_s)$ at $t_s$ (See \autoref{subsec:data}).\\
Use ME-GP to learn the time-evolution of $\Delta_m$ and/or $\Delta_k$ (See \autoref{subsec:me-gp}).\\
 Obtain $\Delta_k(t^*)$ and/or $\Delta_m(t^*)$ at $t^*$, $t^* > \tau$ (See \autoref{subsec:me-gp}). \\
Substitute $k^* = (1 + \Delta_k(t^*))$ and/or $m^* = (1 + \Delta_m(t^*))$ into the nominal model and solve it to obtain responses expected in the future. \\
Take engineering decision. \\
Repeat steps (2) -- (6) as more data becomes available
\end{algorithm}

The proposed digital twin framework has multiple advantages.
\begin{itemize}
    \item The framework proposed utilizes both physics-driven model (ordinary and partial differential equations) and data-driven models (ME-GP). The physics-driven model ensures extrapolatibility of the proposed digital twin. On the other hand, the data-driven model ensures that the proposed digital twin is not limited by the facts that there may be missing physics.
    \item Including physics-based model also enables us to predict other responses of interest. For example, although we only have sensor information about the damped natural frequency of the system, the proposed digital twin can easily predict other responses such as strains, displacements and velocity.
    \item The fact that we utilize ME-GP enables the digital twin to track even multi-timescale dynamical systems such as the one considered in this paper.
\end{itemize}

\section{Illustration of the proposed framework}
\label{sec:illus}
In this section, we illustrate the performance, utility
and applicability
of the
proposed digital twin framework for multi-timescale dynamical systems.
More specifically, we present results for the problem defined in \autoref{sec:ps}.
Three cases as defined in \autoref{sec:dt_ms}
are considered. 
As already stated, it is assumed that we have access
to the damped natural frequency of the system at different (slow) time-steps and the objective is to learn the 
time-evolution of the mass and/or the stiffness.
Once the time-evolution of the mass and/or stiffness is known, the same can be used to 
determine the response of interest at a given time-instant by solving the physics-driven nominal model.
We illustrate how the proposed digital twin can be used to learn the parameters (mass and stiffness) in the past (interpolation) as well as in the future (extrapolation).
Lastly, to illustrate the difficulty posed by a multi-timescale dynamical system and to showcase the necessity of the ME-GP, we compare the results
obtained using the proposed digital twin with those
obtained using a GP based digital twin \cite{chakraborty2020role}.
\subsection{Digital twin via stiffness evolution}\label{subsec: dt_stiff}
First, we consider the case where the variation in the
collected natural frequency is due to the degradation
in the stiffness of the system.
The time-evolution of stiffness is assumed to be of multi-scale nature as shown in Figure \ref{fig:degrad} (b) (the multi-scale one).
However, neither the pattern of the time-evolution nor the number of scales present in the data is known to us a-priori.
Without loss of generality, the sensor data is assumed to be transmitted intermittently at a certain 
regular time-interval.
To simulate a realistic scenario, the damped natural 
frequency data is contaminated with white Gaussian noise
having a standard deviation $\sigma_0$.
It is to be noted that the frequency of data availability depends on a number of factors including 
the bandwidth of the transmission system and 
cost of data collection.
Therefore, we as engineers are not only interested 
in the behavior of the system at a future time, but also in the behaviour at an intermediate time.

Consider that we have $N_s$ observations of the damped natural frequency $\lambda_s \left(t_s\right)$ equally spaced
in time $t_s \in \left[0,\tau\right]$.
Within the proposed digital twin framework, $\lambda_s \left( t_s \right)$ is processed
in accordance to the procedure described in \autoref{subsubsec:stiff_degrad} to obtain the change in the stiffness $\Delta_{\tilde k} \left(t_s\right)$.
After that, using $t_s$ as the input data and $\Delta_{\tilde k}\left(t_s\right)$ as the output data,
the ME-GP model is trained by using SMC and expectation maximization by using Algorithms \ref{alg:smc} and \ref{alg:me-gp}.
Since the number of scales present in the data is supposed to be unknown for a real-life problem, we have used four experts (different from the two scales that are actually present) within the ME-GP framework.
Automatic relevance determination based Matern covariance function and quadratic mean function is considered for all the experts.
The threshold parameter $C$ is considered to be 0.85 \cite{Chakraborty2018efficient,koutsourelakis2011scalable} and we generate $1000$ samples by using
the SMC sampler.
The trained ME-GP model is used as a surrogate to the 
unknown degradation process.
The stiffness at a given time $t^*$ can be calculated by using \autoref{eq:stiff_vary1} where $\Delta_k\left(t_s\right) \approx \Delta_{\tilde k} (t^*)$ is obtained by using the trained ME-GP model. 
By substituting the estimated stiffness into the 
nominal model in \autoref{eq:eom}, it is possible to predict any response of interest at $t^*$.
Also as more data becomes available, the ME-GP model gets updated.

\begin{figure}[ht!]
    \centering
    \includegraphics[width = 0.6\textwidth]{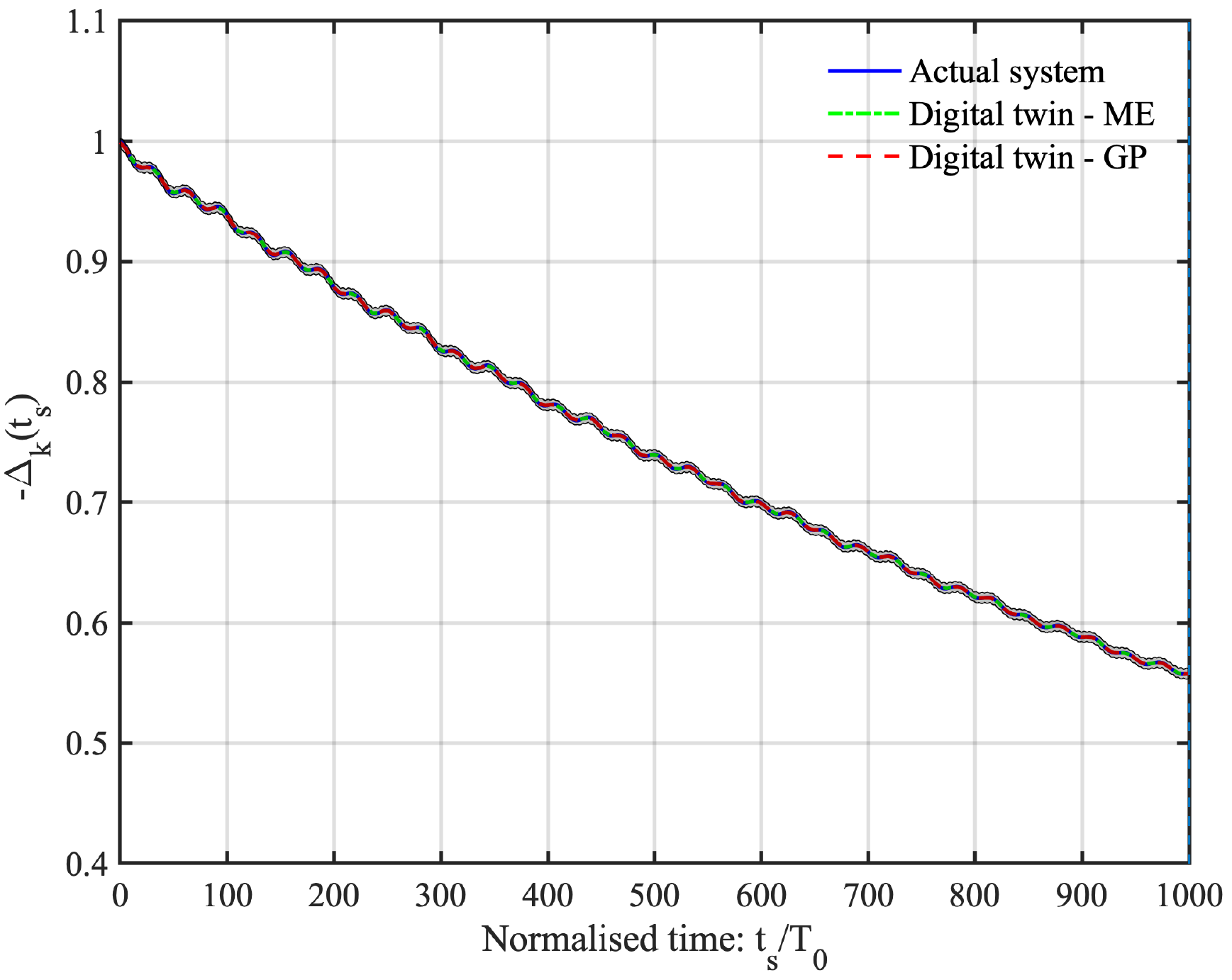}
    \caption{Results obtained using GP based and ME-GP based digital twin for the case where only stiffness changes. Ideal scenario is considered where clean data throughout the service life is available.}
    \label{fig:stiff_perf}
\end{figure}

Figure \ref{fig:stiff_perf} shows the variation of $\Delta_k$ with normalized time.
we have 200 measurements equally spaced throughout the service life of the system.
The measured data is clean (i.e., no noise).
We see that for this case, the proposed digital twin and the GP based digital twin
yields identical results. 
In other words, in the presence of enough (clean) data 
throughout the service life of the system, the proposed ME-GP based digital 
twin reduces to a simple GP based digital twin.
However, in a real-life setting, seldom do we have access to life-time data. Also, the data obtained 
is almost always corrupted by some form of noise.

\begin{figure}[t]
    \centering
    \subfigure[$\tau = 150$]{
    \includegraphics[width = 0.48\textwidth]{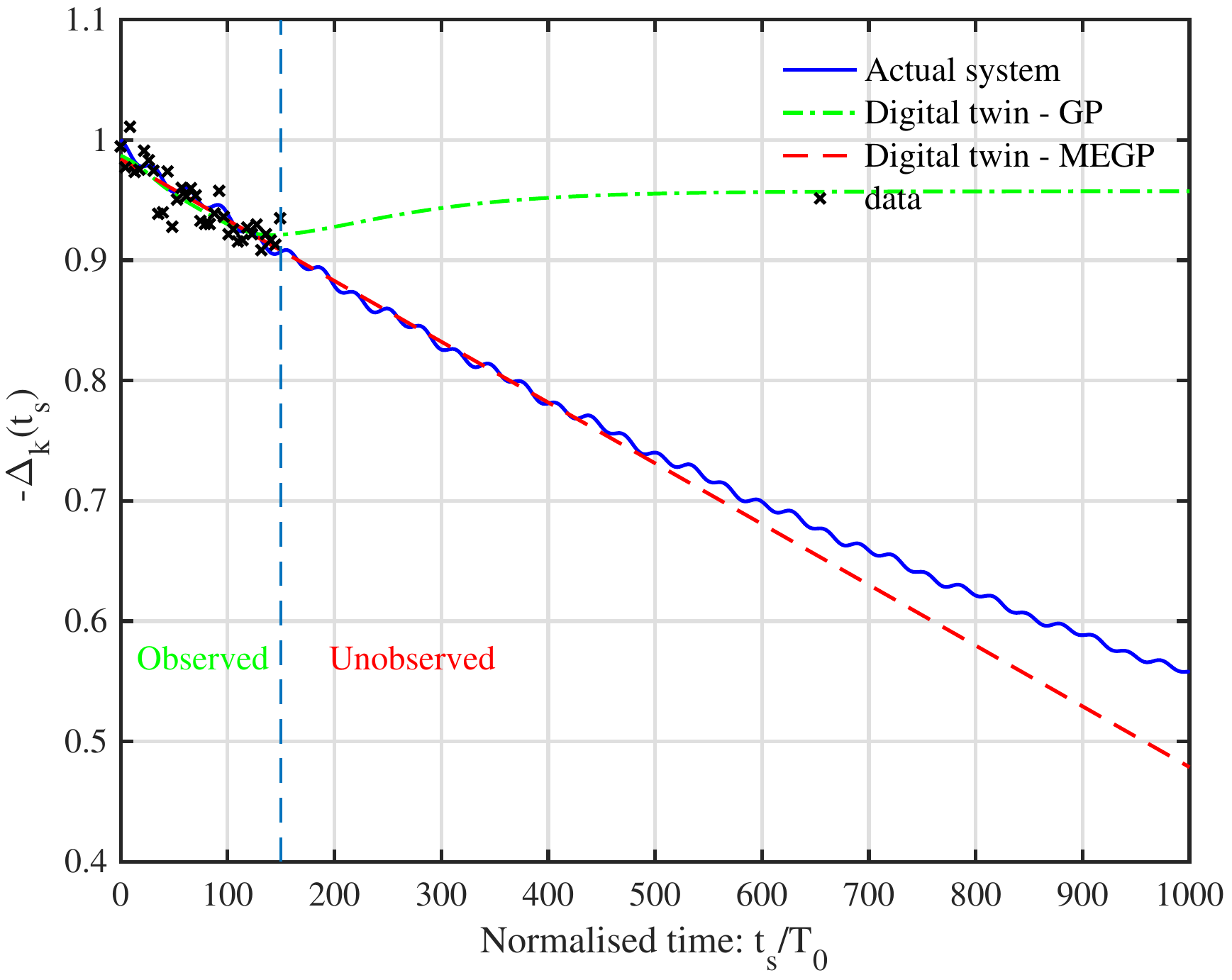}}
    \subfigure[$\tau = 250$]{
    \includegraphics[width=0.48\textwidth]{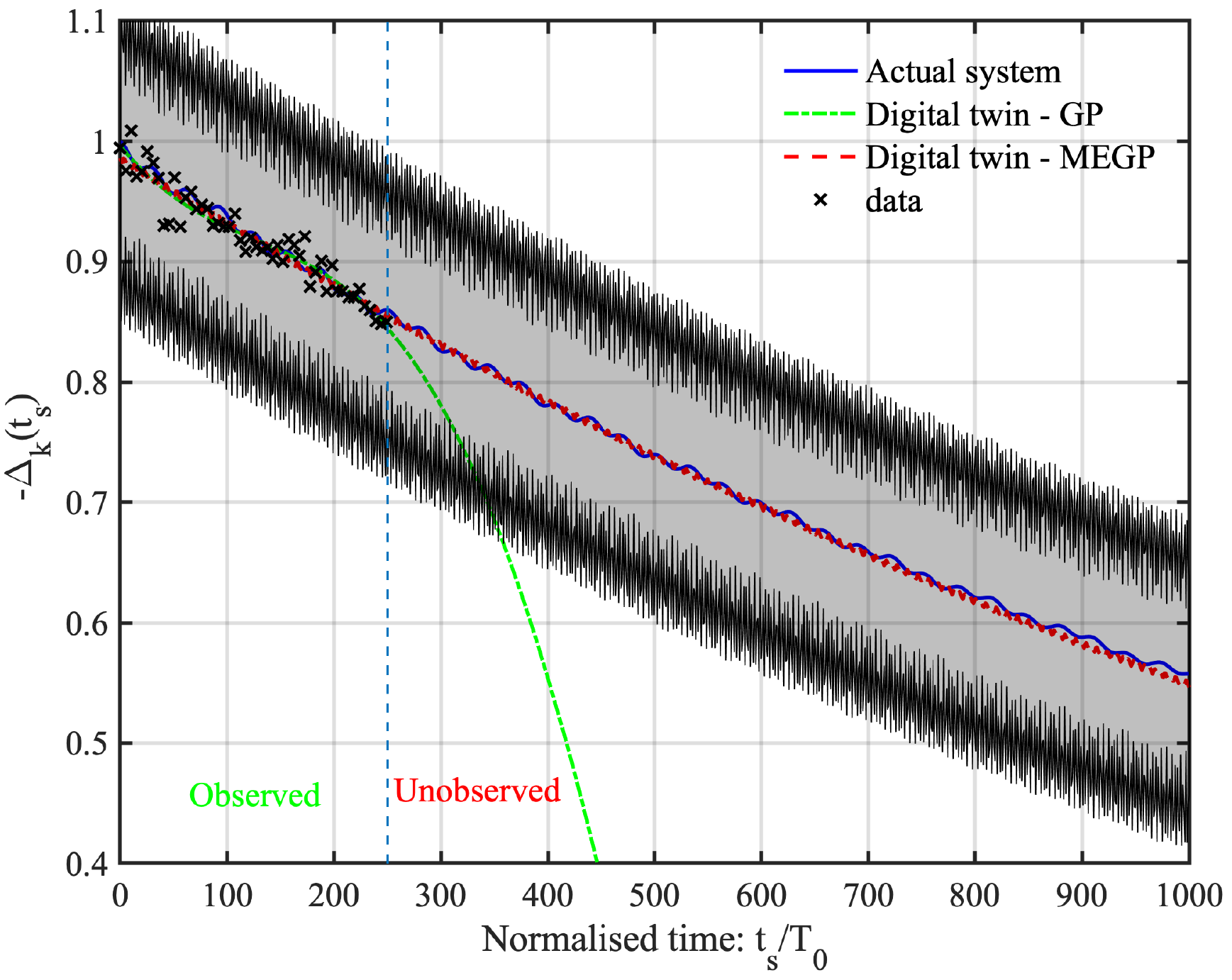}}
    \subfigure[$\tau = 550$]{
    \includegraphics[width=0.48\textwidth]{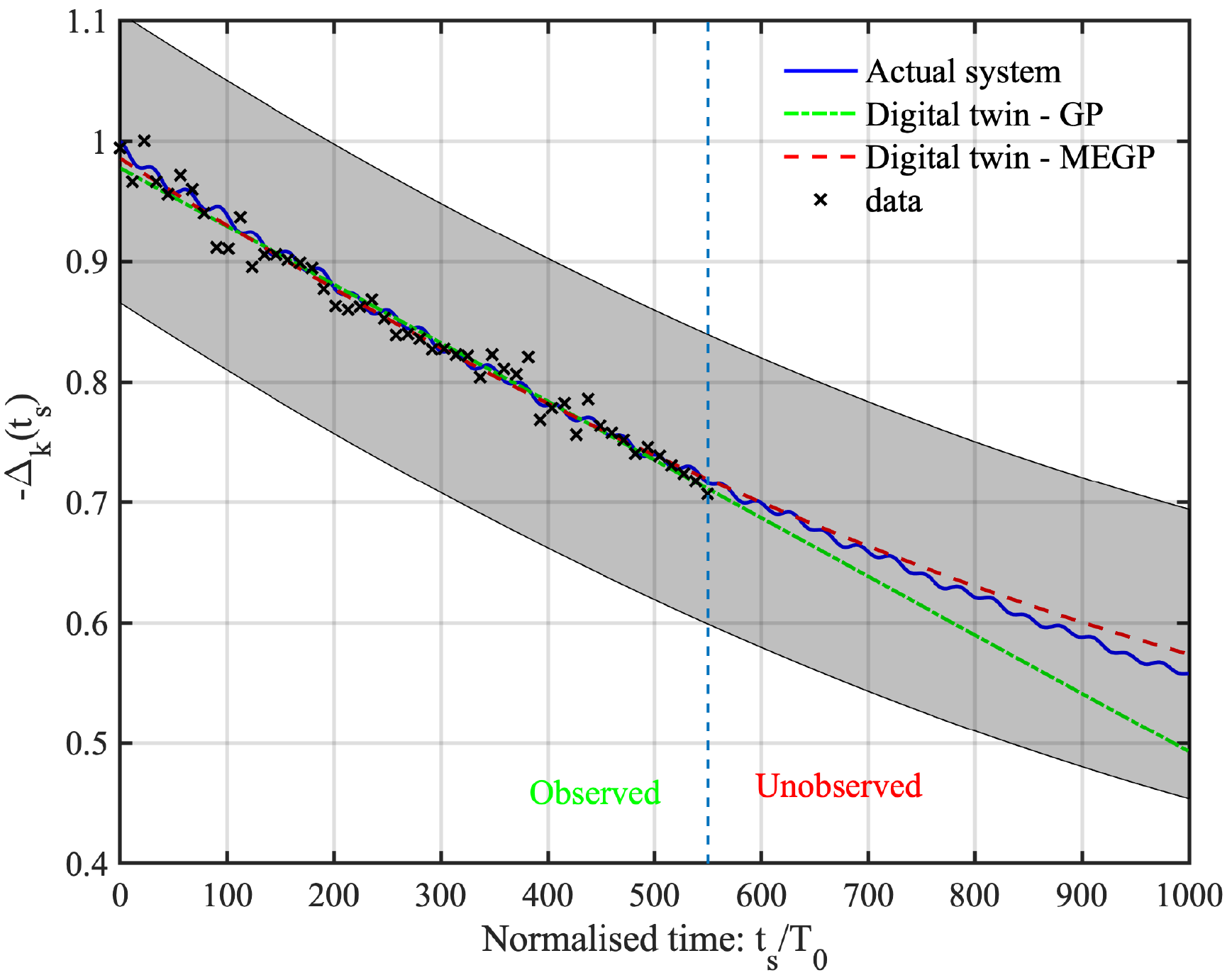}}
    \caption{Results obtained using GP and ME-GP based digital twin with observations up to $\tau = [150, 250, 550]$ and noise variance $\sigma_0 = 0.005$. For $\tau = 150$, we have 35 sensor data whereas for $\tau = [250, 550]$, we have 50 sensor data (shown using cross). The observed and unobserved regime are differentiated by using a vertical line.}
    \label{fig:stiff_005}
\end{figure}

Next, more realistic cases are considered.
More specifically, we consider cases where we only have measurements in a certain observation time-window,
$[0,\tau]$
and we are interested in predicting the evolution of stiffness
at a time $t^*$ where $t^* > \tau$.
Moreover the data collected from the physical system
is considered to be noisy.
\autoref{fig:stiff_005} shows the performance of digital twin for cases where 
$\tau = \left[ 150, 250, 550 \right]$.
For $\tau = 150$, we have 35 measurement data whereas for the other two cases, we have 50 sensor measurements.
For all the three cases, the sensor measurements are contaminated by white Gaussian noise with standard deviation $\sigma_0 = 0.005$.
The observed and unobserved regime are also marked in the figure by using a vertical line.
Similar to \autoref{fig:stiff_perf}, results have been generated by using 
the proposed digital twin
and the GP based digital twin.
For $\tau = 150$ (\autoref{fig:stiff_perf}(a)), the proposed ME-GP based digital twin
is found to yield excellent result up to $t_s/T_0 \approx 600$, which is almost four times the observation time window.
Even beyond $t_s/T_0 = 600$, the results obtained from the proposed digital twin is found to be satisfactory.
The GP based digital twin, on the other hand, is found to yield erroneous result almost immediately after the observation time-window.

One of the necessary characteristic of digital twin
is its capability to update itself as more data becomes available.
Figs. \ref{fig:stiff_perf}(b) and \ref{fig:stiff_perf}(c) show the results when
more data is available and the 
the digital twin has been updated accordingly.
For \autoref{fig:stiff_perf}(b), we have considered that we have access to 50 data points equally spaced in the time-window $[0,250]$.
Similarly, in \autoref{fig:stiff_perf}(c), we have access to 50 data equally spaced in the time-window $[0,550]$.
We see that with more data, the digital twin is able to capture the evolution of stiffness up to $t_s/T_0 = 1000$ (which is assumed to be the service life of the system).
We also see that the predictive uncertainty
for both these cases envelopes all the observed data points.
This indicates that the uncertainty in the system due to limited and noisy measurements have been adequately
captured.
The GP-based digital twin has also been updated for the two cases.
Although the GP based digital twin yields erroneous result for $\tau = 250$, it is found to yield satisfactory results for $\tau = 550$.
However, the ME-GP based digital twin predicted results are still superior.

\begin{figure}[t]
    \centering
    \subfigure[$\tau = 150$]{
    \includegraphics[width = 0.43\textwidth]{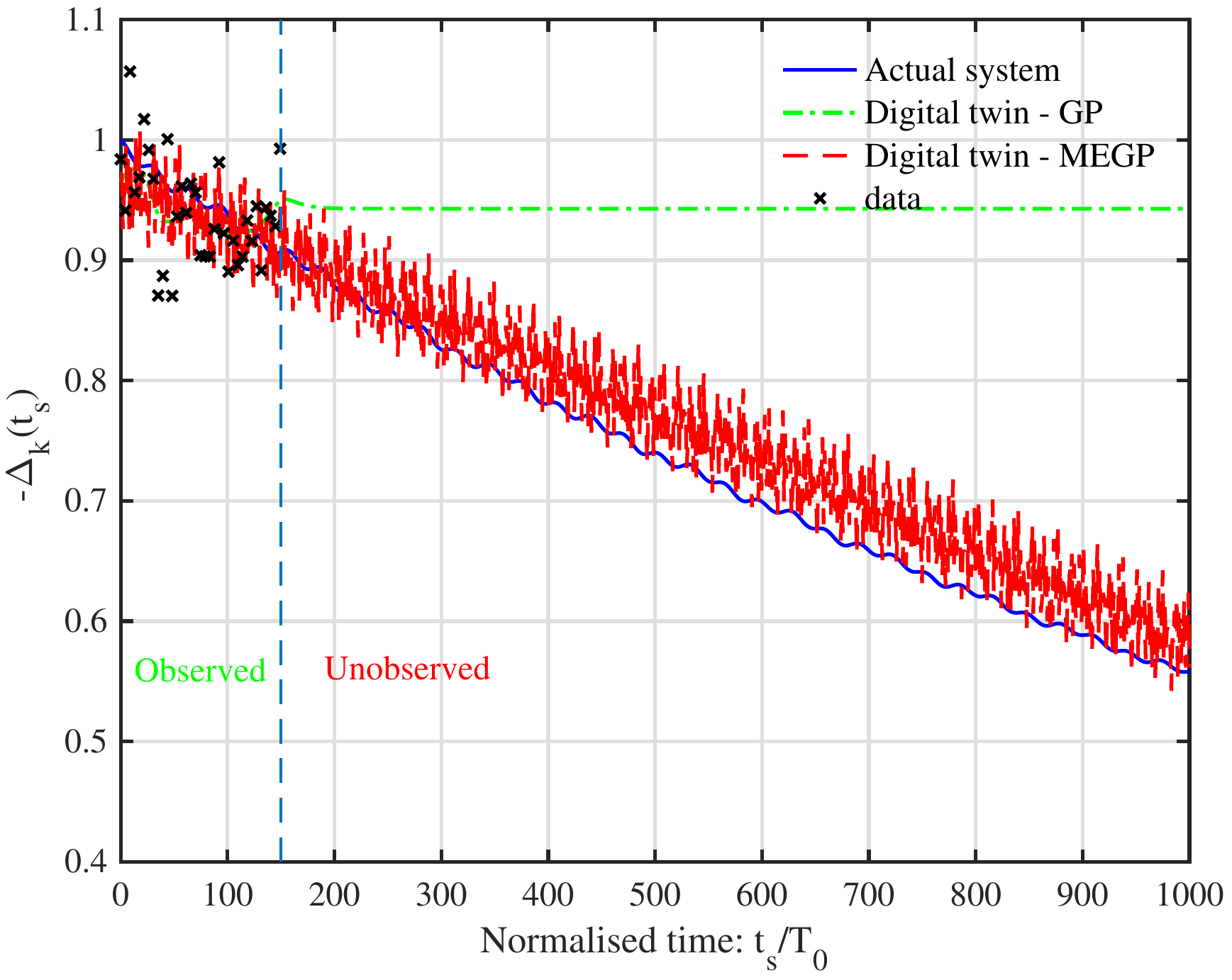}}
    \subfigure[$\tau = 250$]{
    \includegraphics[width=0.48\textwidth]{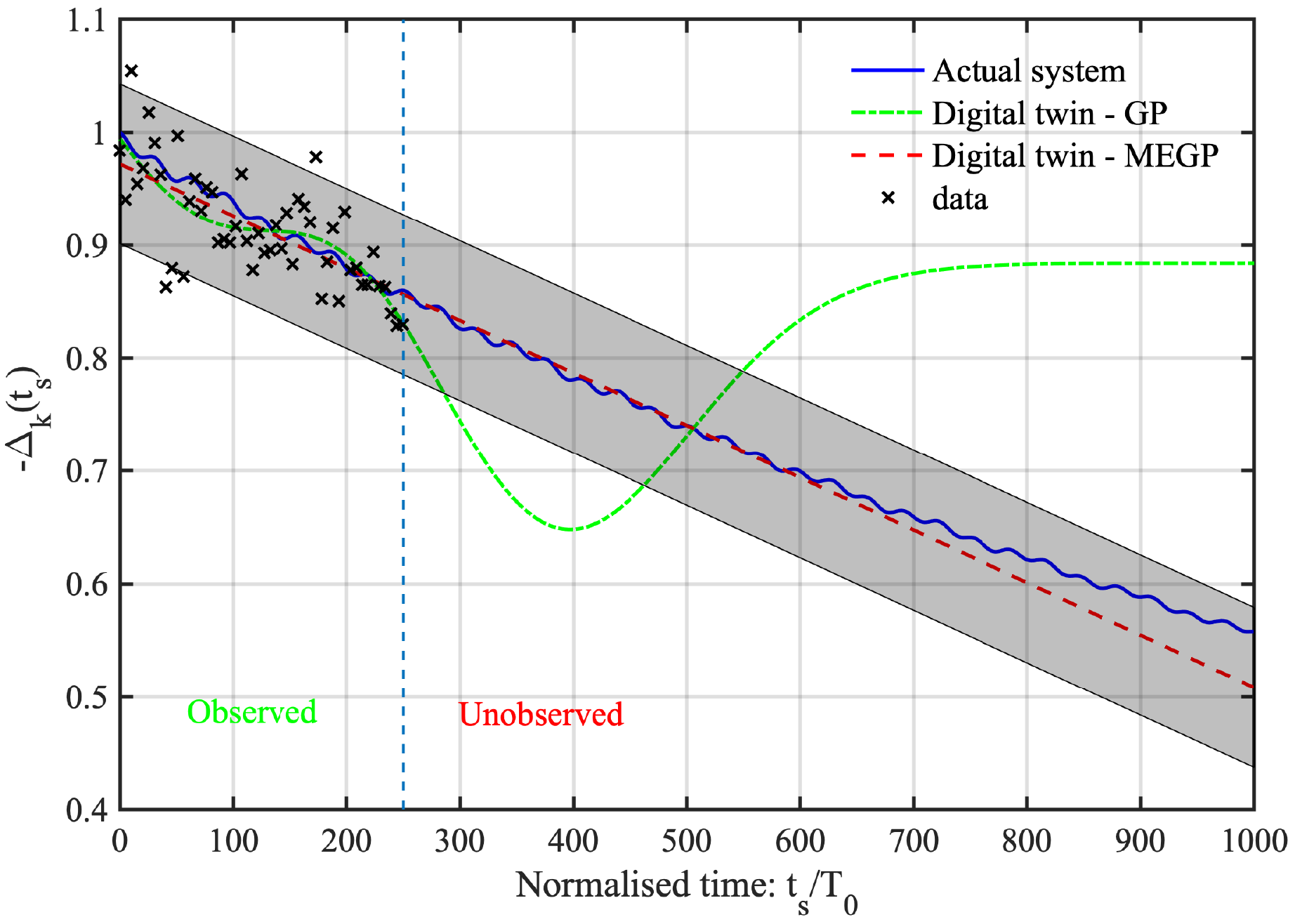}}
    \subfigure[$\tau = 550$]{
    \includegraphics[width=0.48\textwidth]{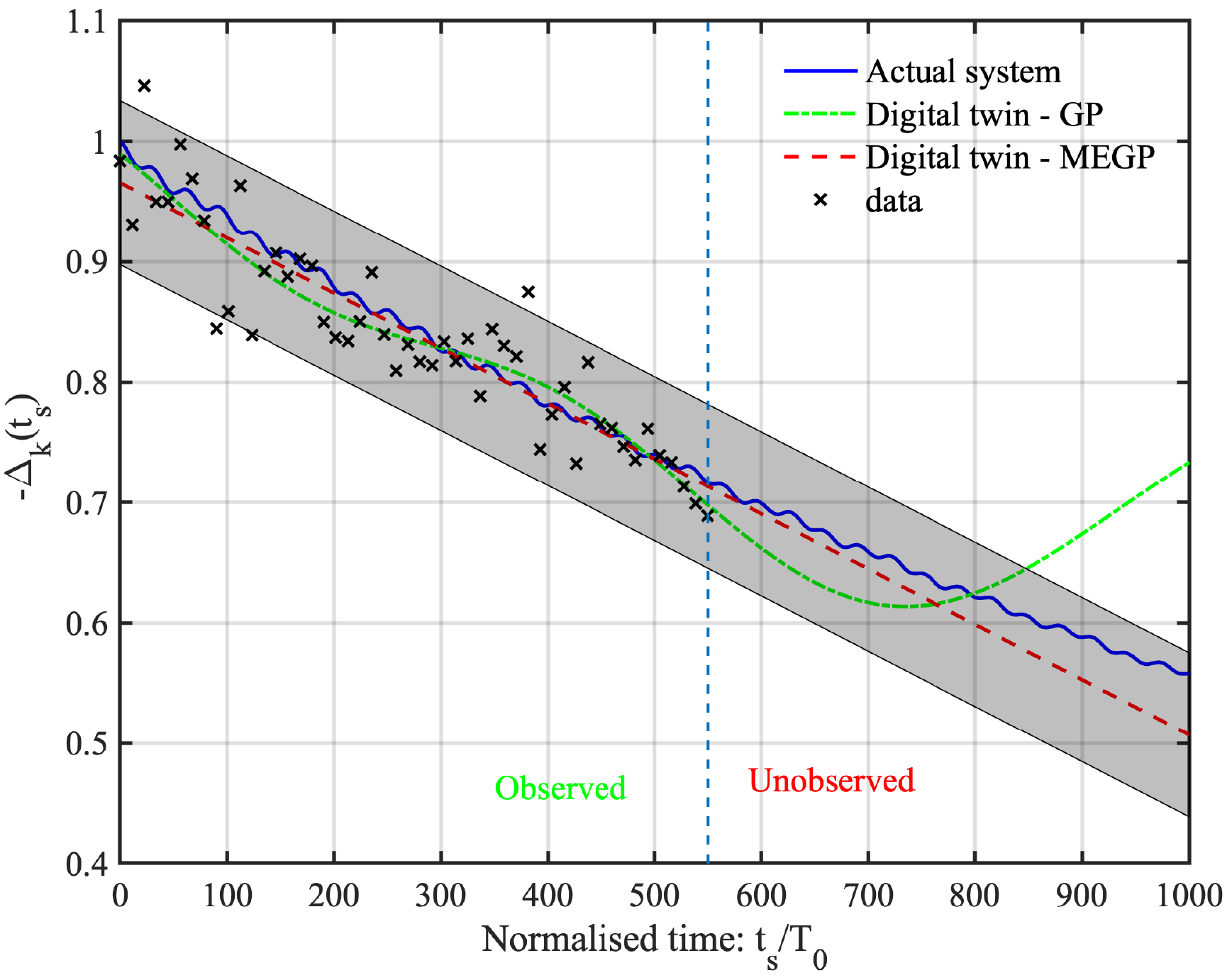}}
    \caption{Results obtained using GP and ME-GP based digital twin with observations up to $\tau = [150, 250, 550]$ and noise variance $\sigma_0 = 0.015$. For observation up to 150, we have 35 sensor data whereas for observation up to 250 and 550, we have 50 sensor data (shown using cross). The observed and unobserved regime are differentiated by using a vertical line.}
    \label{fig:stiff_015}
\end{figure}

Lastly, we investigate the case when the noise variance is more. 
\autoref{fig:stiff_015} shows the results corresponding to $\sigma_0 = 0.015$.
Same three cases as \autoref{fig:stiff_005} are considered.
We observe that for $\tau = 150$, the prediction obtained using ME-GP based digital twin oscillates around the actual solution (\autoref{fig:stiff_015} (a)).
This oscillatory behaviour is probably because ME-GP overfits to the noise in the observations. The GP based digital twin is found to yield erroneous results.
The digital twin models are then updated by collecting data up to $\tau = 250$ and $\tau = 550$.
For both these cases, ME-GP based digital twin is found to yield reasonably good predictions. 
However, because of the increase in the noise variance, 
the predicted results are inferior to those shown in
\autoref{fig:stiff_005}.
This indicates the importance of collecting clean data from the physical system.
The GP based digital twin for all the three cases in \autoref{fig:stiff_015} are found to yield erroneous results beyond the observation window.
This indicates the superiority of the proposed ME-GP over GP, particularly when predicting the future.

\subsection{Digital twin via mass evolution}\label{subsec:dt_mass}
Next, we consider the case corresponding to \autoref{subsubsec:mass_degrad} where
the variation in the damped natural frequency of the system is due to the variation in
its mass. 
The time-evolution of mass is considered to be of multi-scale nature as shown in \autoref{fig:degrad}(a).
However, similar to the stiffness degradation case,
neither the pattern of time-evolution nor the number of scales present in the data is known to us a-priori.
We again assume the sensor data to be transmitted intermittently at a fixed time interval.
To emulate a realistic scenario, the damped natural frequency data is contaminated with white Gaussian noise having a standard deviation $\sigma_0$.
The objective is to use the proposed digital twin to
predict the multi-scale time evolution of mass.

Consider that we have $N_s$ observations of the damped natural frequency $\lambda_s (t_s)$ equally spaced in time 
$t_s \in \left[0, \tau \right]$.
Using the procedure described in \autoref{subsubsec:mass_degrad}, we first process $\lambda_s (t_s)$ to obtain the change in the mass
$\Delta_{\tilde m} \left(t_s \right)$.
Thereafter, using $t_s$ as the input and $\Delta_{\tilde m} \left(t_s\right)$ as the output,
we train an ME-GP model by using Algorithms \ref{alg:smc} and \ref{alg:me-gp}.
The setup for the algorithms are kept similar to that described in \autoref{subsec: dt_stiff}
and the trained ME-GP model is treated as a surrogate
of the unknown $\Delta_m$.
For obtaining the response of the system at $t^*$, one can  utilize the trained ME-GP model to obtain the 
updated $m$ at $t^*$ and then substitute it 
into the nominal model and solve it to obtain the
responses of interest.
As more data becomes available, the digital twin is updated.

\begin{figure}[ht!]
    \centering
    \includegraphics[width=0.7\textwidth]{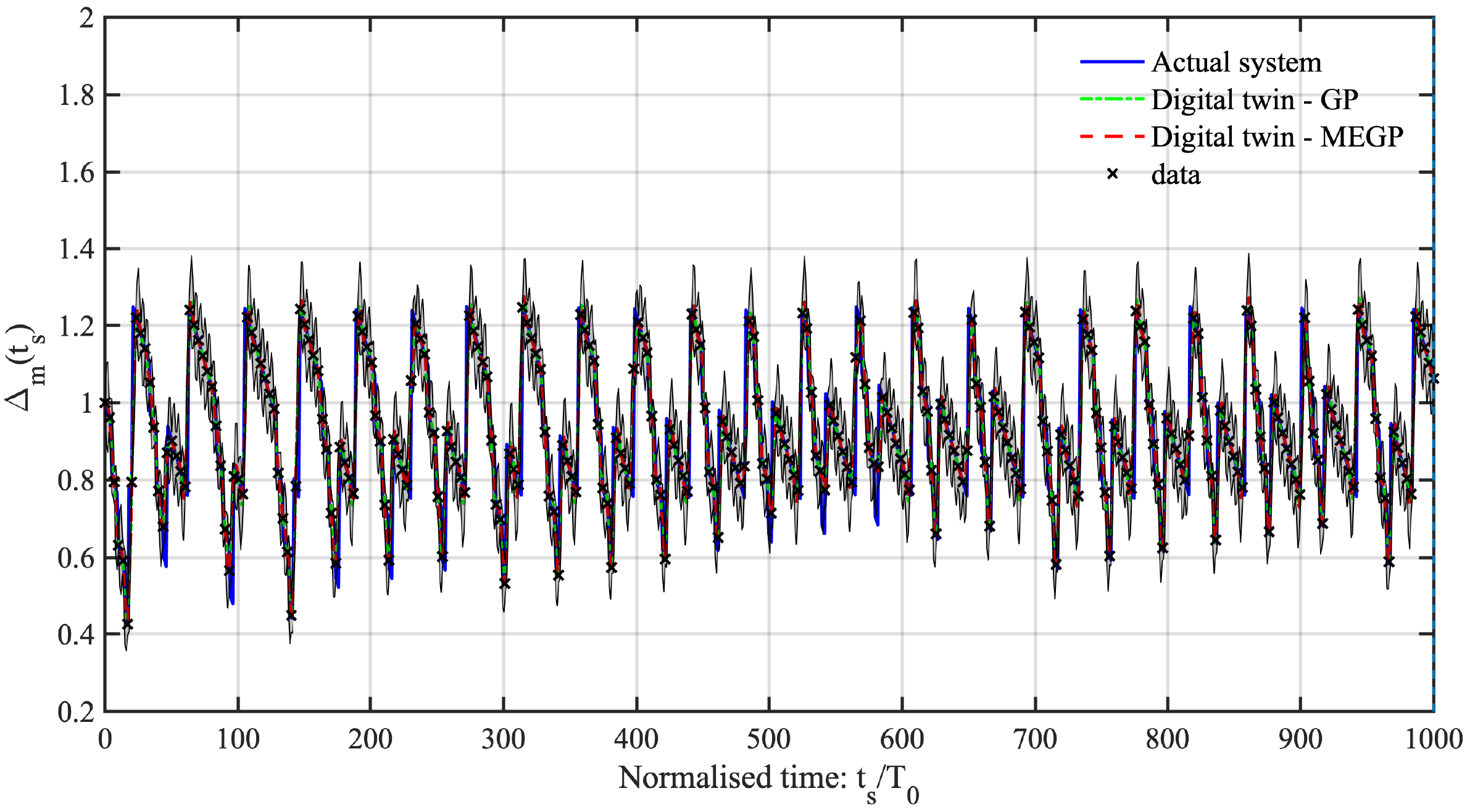}
    \caption{Results obtained using GP based and ME-GP based digital twin for the case where only mass changes. Ideal scenario is considered where equally spaced clean data throughout the service life is available.}
    \label{fig:mass_perf}
\end{figure}

Figure \ref{fig:mass_perf} shows the variation of $\Delta_m$ with normalized time $t_s/T_0$.
These results correspond to the perfect case where 
we have 300 measurements equally spaced throughout the
service life of the system.
The data available are also free of any noise.
We observe that the proposed digital twin and the
GP based digital twin yields identical results.
In other words, for this case, the proposed ME-GP based digital twin reduces to the GP based digital twin
proposed in \cite{chakraborty2020role}.
However, one must note that this is an ideal scenario.
In a more realistic setting, the data collected will be noisy and, almost always, data over the complete service-life will not be available.

\begin{figure}[ht!]
    \centering
    \subfigure[$\tau = 150$]{
    \includegraphics[width = 0.75 \textwidth]{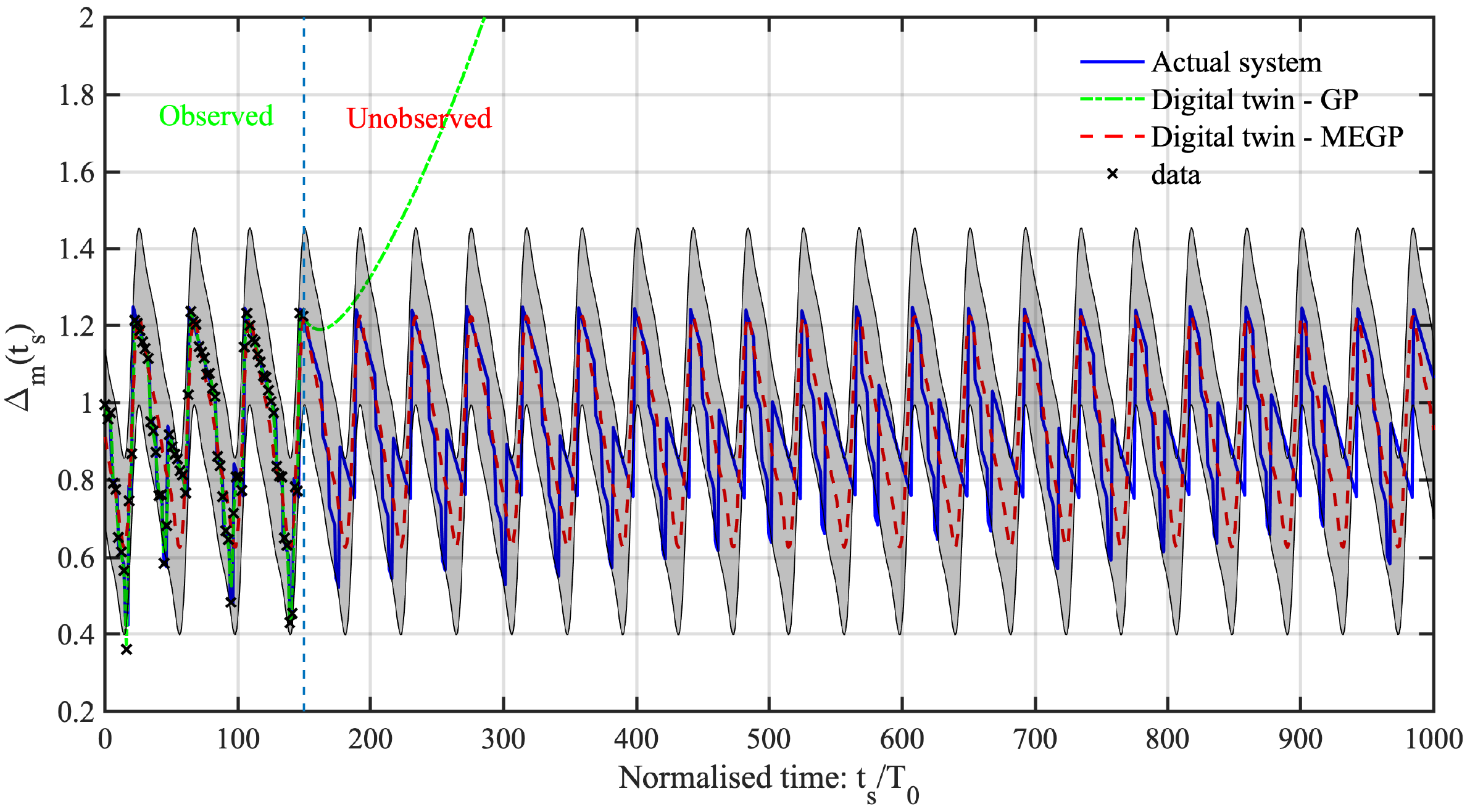}}
    \subfigure[$\tau = 550$]{
    \includegraphics[width = 0.75 \textwidth]{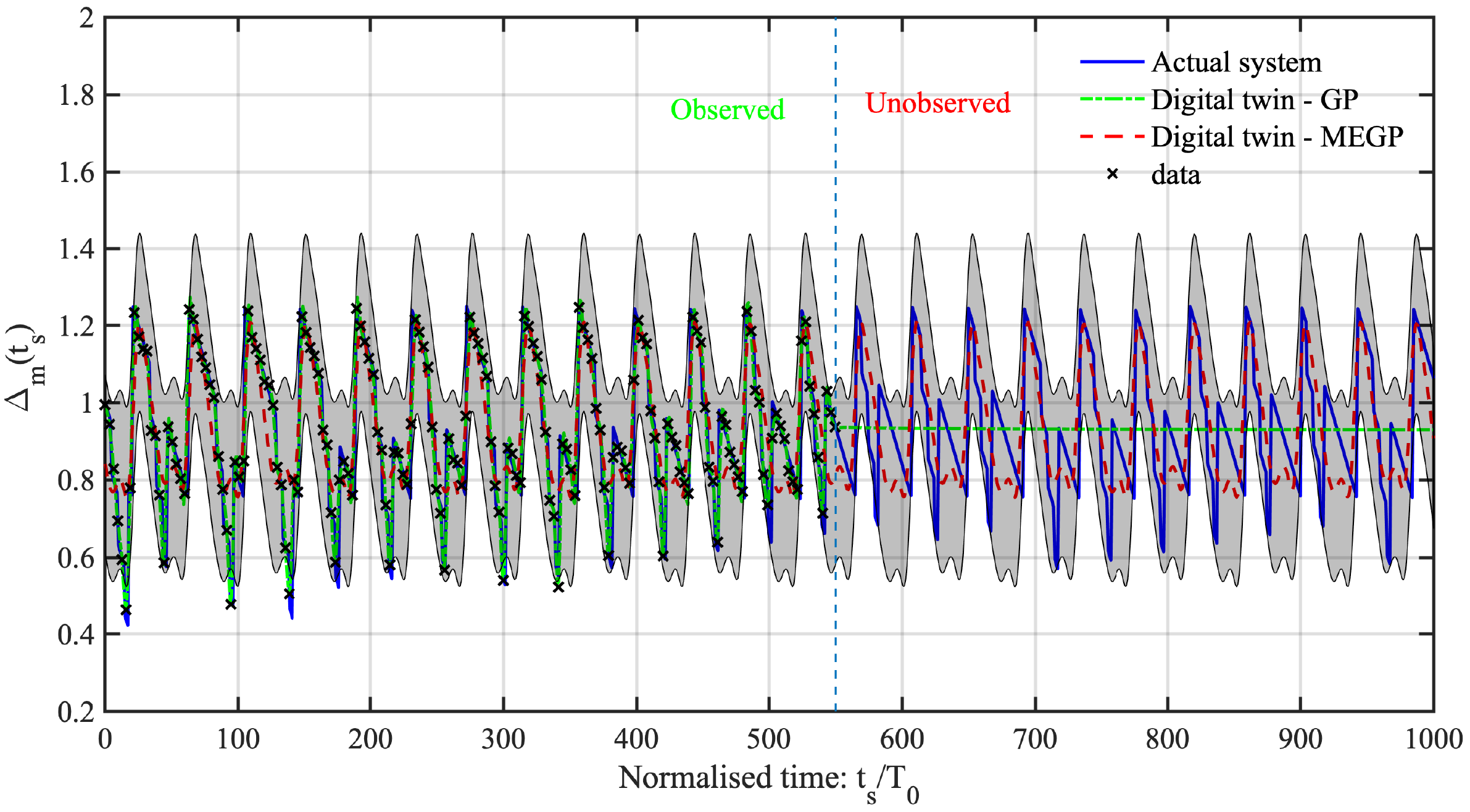}}
    \caption{Results obtained using GP and ME-GP based digital twin with observation window of [0, 150] and [0, 550], and $\sigma_0$= 0.005.  For observation window of [0, 150], we have 75 sensor data whereas for observation window of [0, 550], we have 175 sensor data (shown using cross). The observed and unobserved regime are differentiated by using a vertical line}
    \label{fig:mass_005}
\end{figure}

Next we consider a more realistic setup where we have data over a certain time-window $[0, \tau]$ and the goal is to predict $\Delta_m$ at a time $t^*$ where $t^* > \tau$.
The data collected is contaminated by white Gaussian 
noise.
\autoref{fig:mass_005}(a) shows the digital twin predicted results corresponding to $\tau = 150$.
The data collected is contaminated by white noise with $\sigma_0 = 0.005$.
The observed and unobserved regimes are marked by a vertical line.
Results using GP based digital twin have also been generated for the sake of comparison.
For $\tau=150$, the ME-GP model trained from only 
75 data reasonable predicts the evolution
of mass over the 
the service life of the system. 
More specifically, the primary crest of the response has been accurately captured throughout the service life. However, as time increases, the digital twin over-predicts the trough.
The proposed approach being Bayesian in nature also provides the predictive uncertainty as indicated 
by the shaded plot.
Throughout the service life, the true solution
is within the shaded plot, indicating that the  uncertainty due to limited and noisy data is properly captured.
The GP based digital twin, although provides excellent prediction in $t_s/T_0 = [0, 150]$, yields erroneous prediction almost immediately beyond the observation time-window.

Figure \ref{fig:mass_005}(b) shows the results obtained using the digital twins 
for $\tau = 550$.
175 equally spaced observations over the time-window are collected.
The data collected is contaminated with white Gaussian noise having $\sigma_0 = 0.005$.
Again, the proposed ME-GP based provides reasonable predictions.
The GP based digital twin, on the other hand, fails 
to capture the time evolution of $\Delta _m$ beyond 
the training time window.

\begin{figure}[ht!]
    \centering
    \subfigure[$\tau = 150$]{
    \includegraphics[width = 0.75 \textwidth]{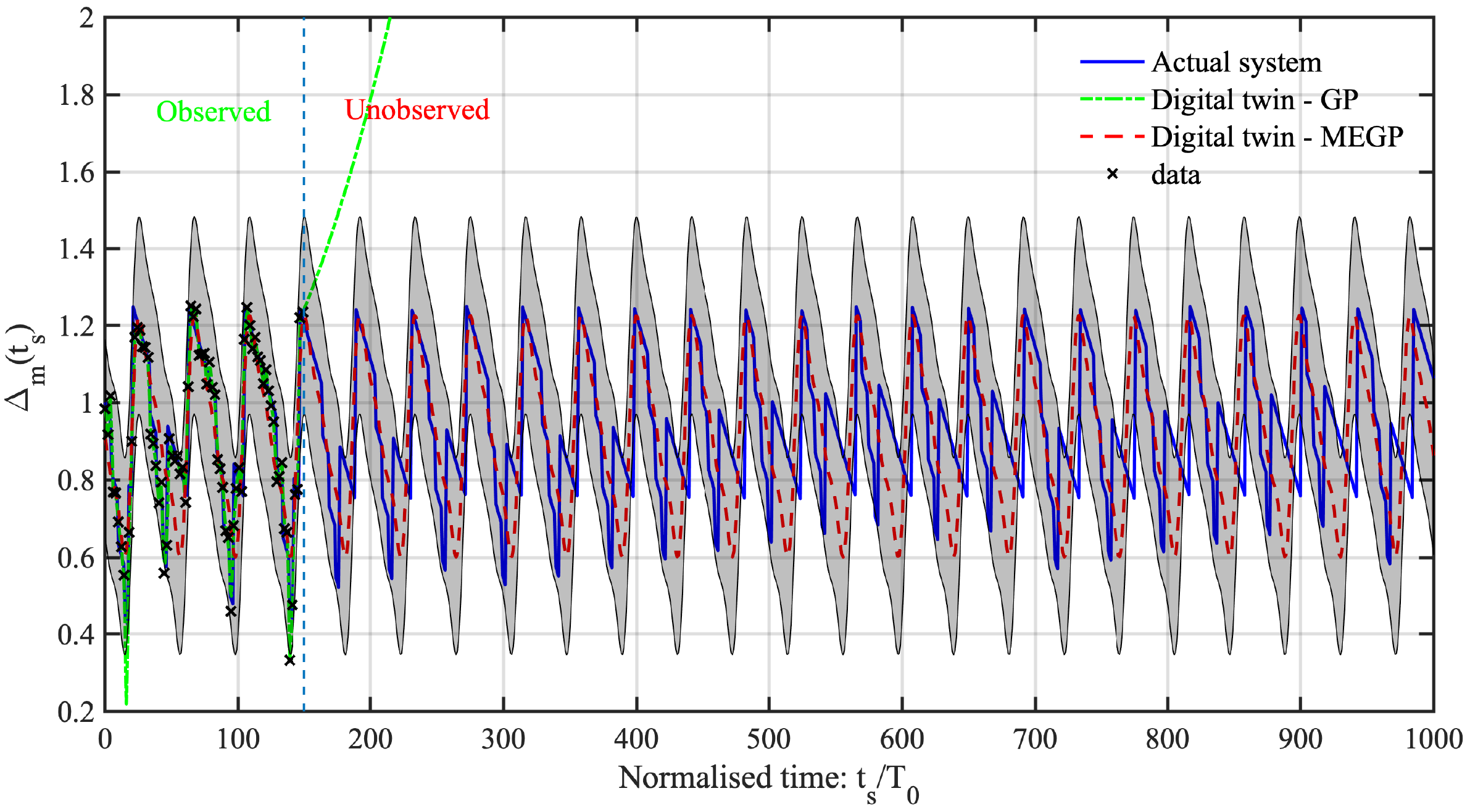}}
    \subfigure[$\tau = 550$]{
    \includegraphics[width = 0.75 \textwidth]{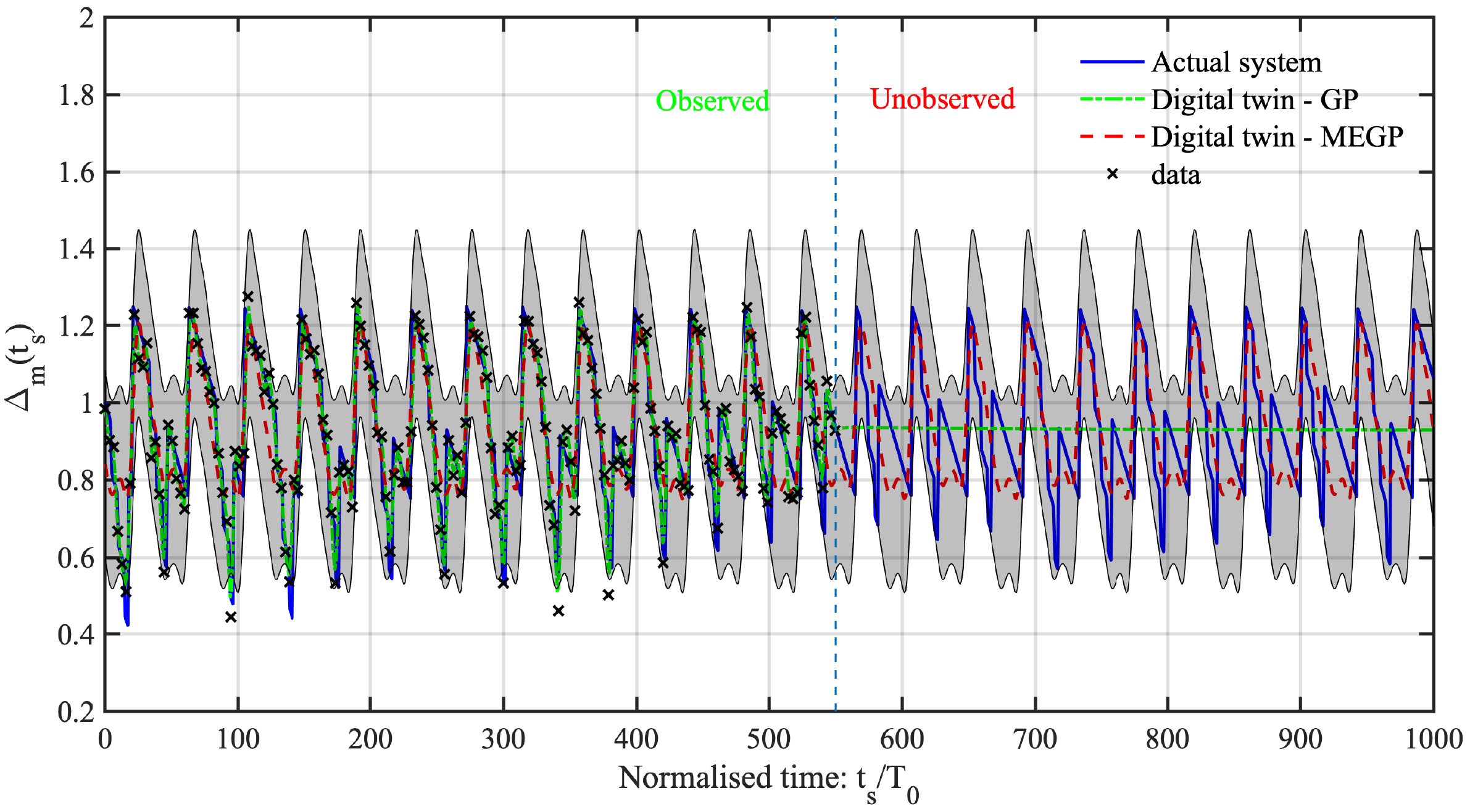}}
    \caption{Results obtained using GP and ME-GP based digital twin with observation window of [0, 150] and [0, 550], and $\sigma_0$= 0.015.  For observation window of [0, 150], we have 75 sensor data whereas for observation window of [0, 550], we have 175 sensor data (shown using cross). The observed and unobserved regime are differentiated by using a vertical line}
    \label{fig:mass_015}
\end{figure}

Lastly, we consider the case when the noise variance
is more.
\autoref{fig:mass_015} shows the results 
corresponding to $\sigma_0 = 0.015$.
The predicted time-evolution of mass is found to be similar to that predicted in \autoref{fig:mass_005}.
The predictive uncertainty is found to increase for this case. 
This can be attributed to the increase in the noise in the data.
GP based digital twin was unable to properly predict the time-evolution beyond the observation window.

\subsection{Digital twin via mass and stiffness evolution}\label{subsec:dt_mass_stiffness}
Lastly, we illustrate the performance of the proposed
digital twin when both mass and stiffness evolves with time.
Time evolution of both mass and stiffness are of 
multi-scale nature as shown in \autoref{fig:degrad}.
Similar to the previous cases, the sensor data is assumed to be transmitted intermittently 
at a fixed time interval.
Considering we have $N_s$ observations of the 
damped natural frequency, the digital twin first process this data to obtain $\Delta_m$ and $\Delta_k$.
Details on the data processing step are furnished in
\autoref{subsubsec:mass_and_stiff_degrad}.
Thereafter, ME-GP is used to learn the time evolution
of mass and stiffness.
The parameters for the ME-GP algorithm are kept identical as that discussed in \autoref{subsec:dt_mass_stiffness}.
For obtaining response at a given time-instant $t^*$,
one first obtains the mass and stiffness by using the trained ME-GP model as a surrogate.
Thereafter, the responses of interest are obtained
by substituting the ME-GP predicted mass and stiffness
into the nominal model and solving it.
Similar to the previous two cases, we only present the performance of the digital in predicting the
time evolution of mass and stiffness; the argument being, if the time evolution of mass and stiffness are accurately captured, the responses predicted will also be accurate.

\begin{figure}[ht!]
    \centering
    \subfigure[$\Delta _m$]{
    \includegraphics[width = 0.7\textwidth]{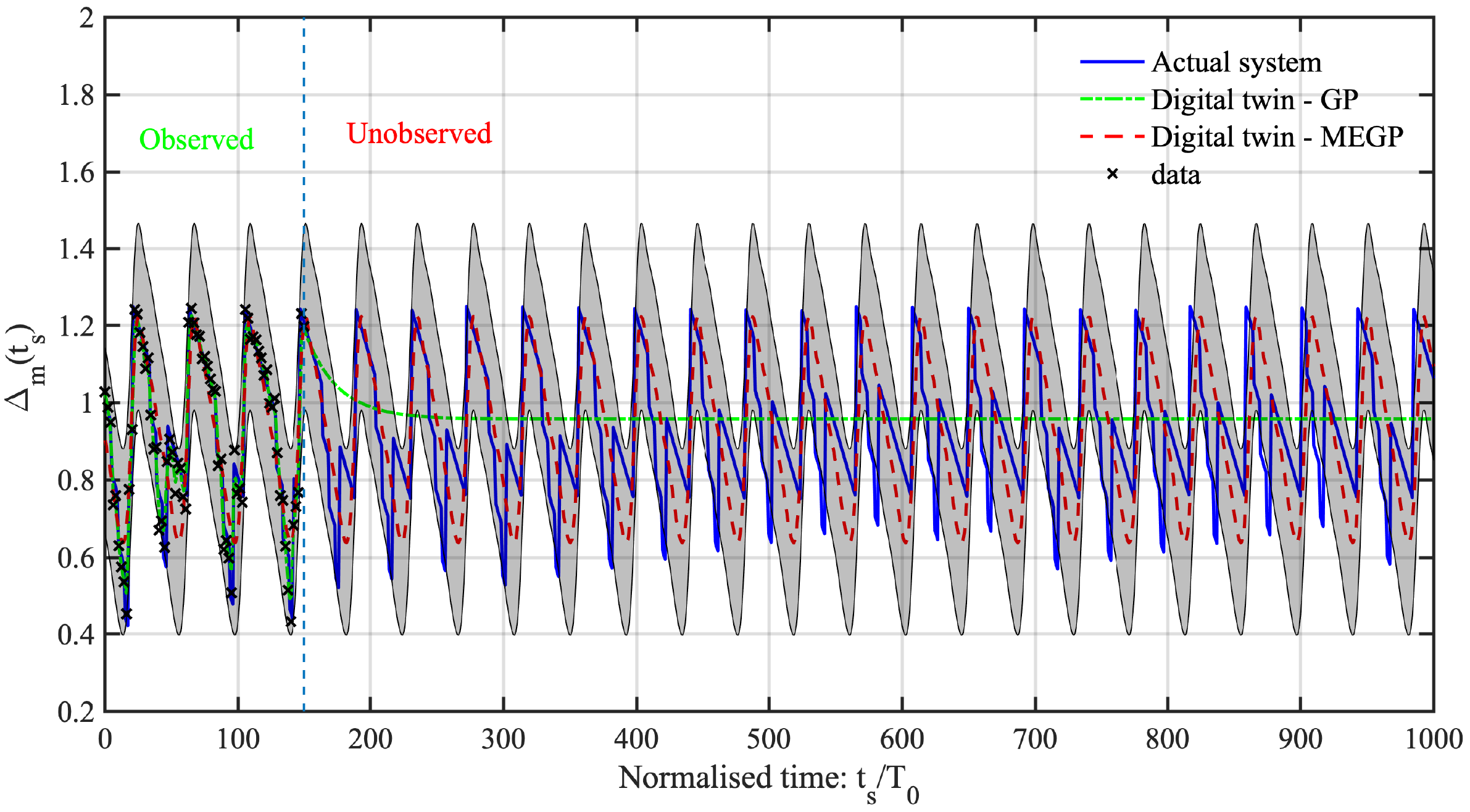}}
    \subfigure[$\Delta _k$]{
    \includegraphics[width = 0.7\textwidth]{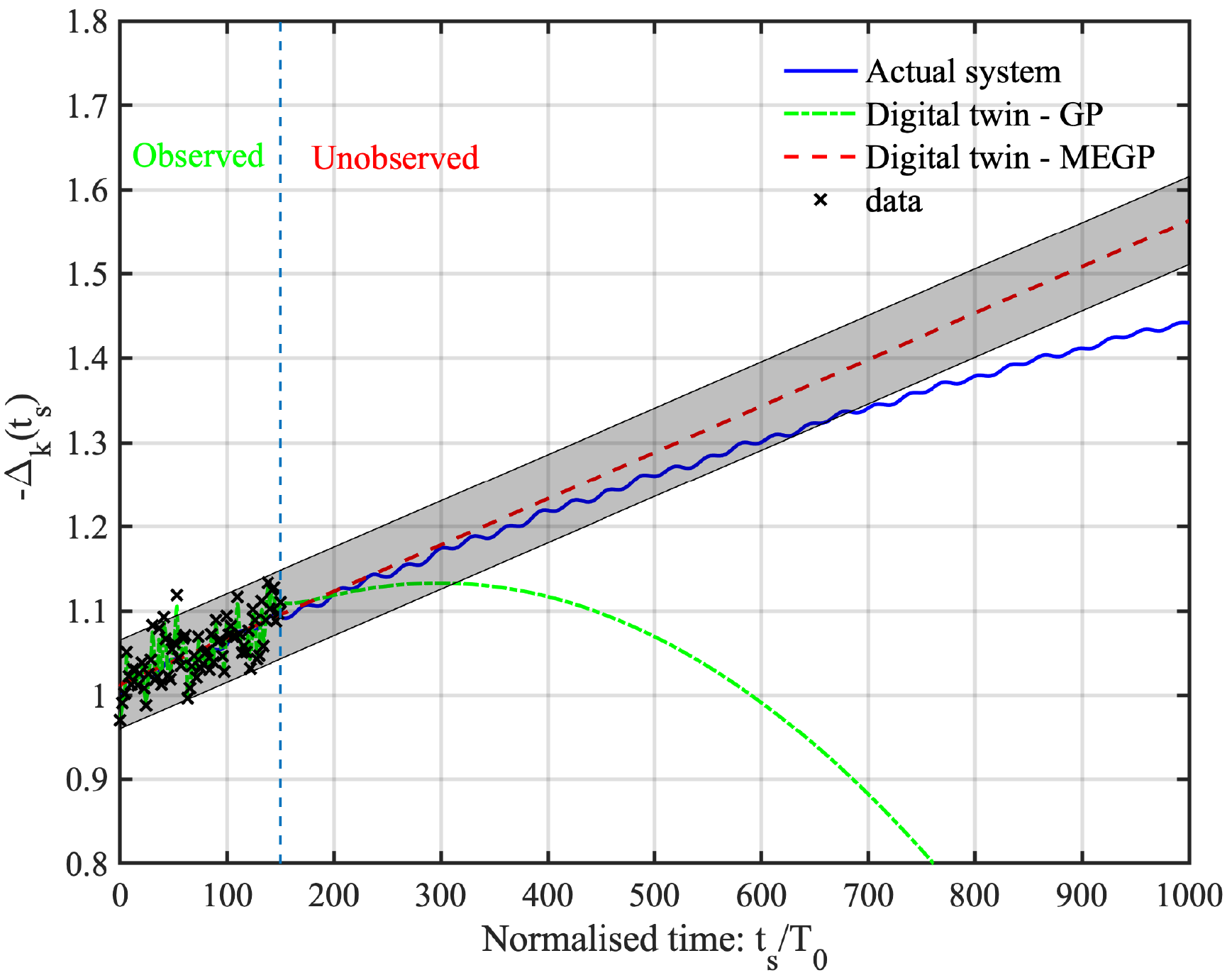}}
    \caption{Digital twin predicted responses, $\Delta_m$ and $\Delta_k$. The digital twin is trained using 75 equally spaced sensor measurements in observation window $[0, 150]$. The sensor data are contaminated by white Gaussian noise with $\sigma_0 = 0.025$. The observed and unobserved regime are shown by a vertical line.}
    \label{fig:mk_025_75}
\end{figure}

Given the fact that the proposed digital twin was found to yield almost exact
result for the ideal scenario for the previous two cases, we directly proceed to the realistic case-studies. 
First, we consider the case where the observation
window is $[0,150]$.
Within this time-window, it is assumed that we have
access to 75 equally spaced sensor measurements.
The data collected is contaminated by a white Gaussian noise with $\sigma_0 = 0.025$.
\autoref{fig:mk_025_75} shows the evolution of $\Delta_m$ and $\Delta_k$
predicted using the digital twin.
For $\Delta_m$, it is observed that the digital twin yields reasonable prediction throughout the service life of the system.
However, for $\Delta_k$, the results are found to deviate beyond $t_s/T_0 = 350$.
Since the proposed approach is Bayesian in nature, 
the predictive uncertainty has also been computed.
For $\Delta_m$, the predictive uncertainty envelopes the true behaviour throughout the service life; this indicates that the uncertainty due to noise and limited data has been appropriately captured.
However, for $\Delta_k$, the true solution is outside the envelope beyond $t_s/T_0 = 600$.
This indicates that the proposed digital twin is over-confident when $t_s/T_0 > 600$.
The GP based digital twin for both $\Delta_m$ an $\Delta_k$ yields erroneous results beyond $t_s/T_0 = 200$.

\begin{figure}[ht!]
    \centering
    \subfigure[$N_s = 120$]{
    \includegraphics[width = 0.7\textwidth]{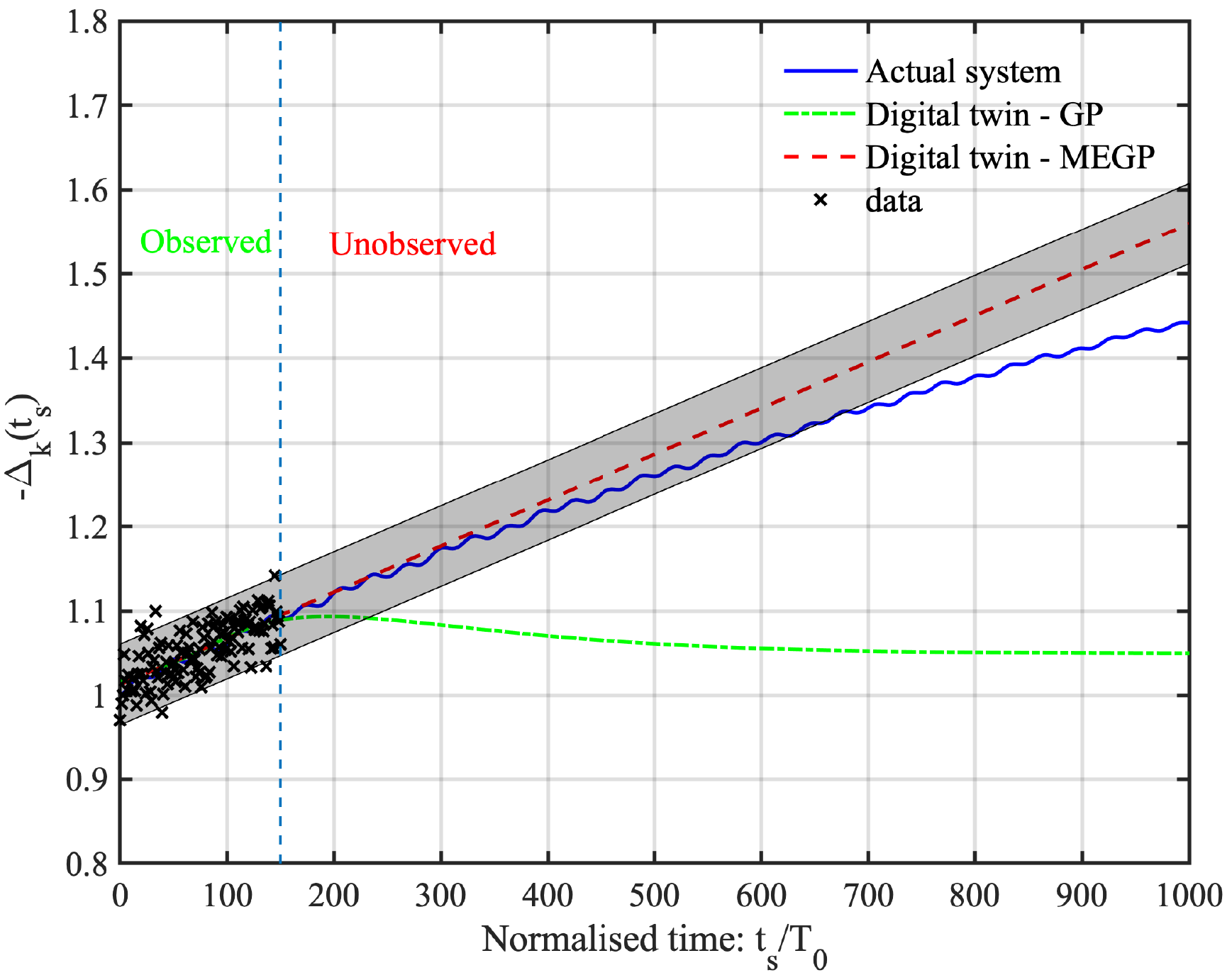}}
    \subfigure[$N_s = 150$]{
    \includegraphics[width = 0.7\textwidth]{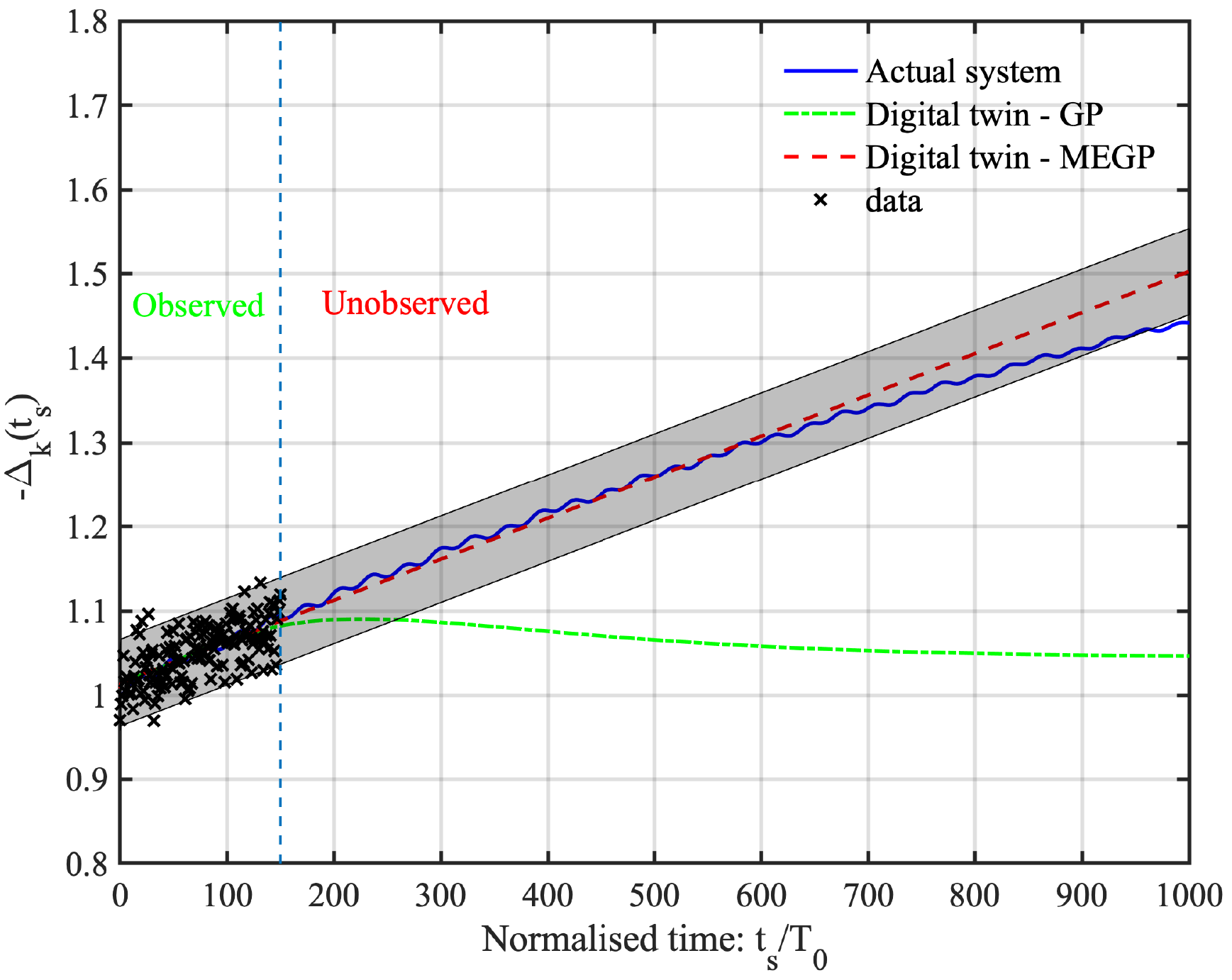}}
    \caption{Digital twin performance with increase in number of sensor measurements over the observation window $[0,150]$. The sensor data are contaminated with white Gaussian noise having $\sigma_0 = 0.025$.}
    \label{fig:mk_025_150}
\end{figure}

\begin{figure}[ht!]
    \centering
    \includegraphics[width = 0.6 \textwidth]{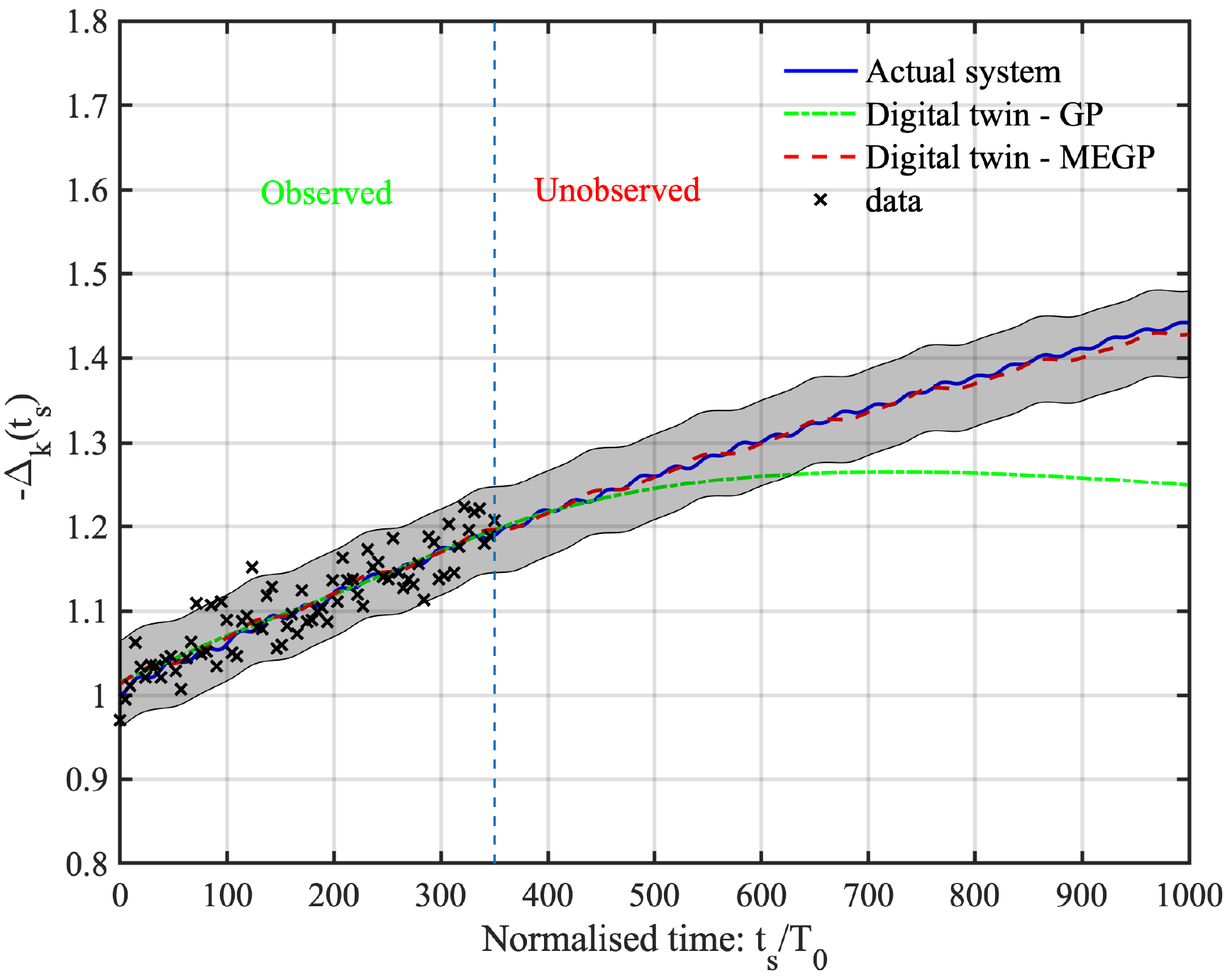}
    \caption{Performance of digital twin for observation time-window $[0,350]$. 75 sensor data are available within the specified time-window. The data are contaminated by white Gaussian noise with $\sigma_0 = 0.025$.}
    \label{fig:mk_35_75}
\end{figure}

For improving the performance of the digital twin,
we carried out investigation by providing additional
sensor data to the model.
More specifically, we provided 120 and 150 equally 
spaced observations within the same observation window $[0, 150]$.
The results are shown in \autoref{fig:mk_025_150}.
As the predictions for $\Delta_m$ was already reasonable in \autoref{fig:mk_35_75}, only the results corresponding to $\Delta_k$ are
presented.
We observe that with an increase in the number of
observations, the digital twin predictions become more and more closer to the actual solution.
The predictive uncertainty is also found to improve as it envelopes the true solution.

As the last case study, we carry out an investigation
by increasing the observation window to $[0,350]$.
However, the number of sensor observations is 
kept fixed at 75.
The result is shown in \autoref{fig:mk_35_75}
With this setup, the digital almost perfectly predicts the time evolution of the stiffness.
This illustrates the importance of collecting data 
over a longer time-span.
For all the cases discussed, the GP based digital twin fails to provide accurate prediction beyond the
observation window.

\section{Discussion}\label{sec:discussions}

The dynamical system observed in engineering and technology often exhibits multiple time-scales. Tracking and solving such systems are challenging due to the need for very fine temporal discretization. In this paper, we propose a digital twin framework for multi-timescale dynamical systems. The key ideas proposed in this work include: 
\begin{itemize}

    \item For the first time, digital twin for multi-timescale dynamical systems have been proposed.    The framework proposed, from very few data, can track the multi-scale evolution of the system parameters and use the same for predicting future responses.        
    
    \item The digital twin proposed fuses a physics-driven nominal model with a data-driven machine learning model. The physics-driven model ensures the generalization of the proposed framework and enables the digital twin in predicting unobserved responses of interest.    
    
    \item The data-driven machine learning model compensates for the fact that the physics of the problem is often not completely defined.    
    In this paper, the machine learning-based model is utilized for learning the time-evolution of the mass and/or stiffness of the system.        
    
    \item As the machine learning model, we propose to use a mixture of experts. We propose to use GP as experts within the mixture of experts framework (see, \autoref{alg:me-gp}). The idea is to let each expert track the time-evolution at one scale. For learning the parameters of the model, an algorithm based on expectation-maximization and sequential Monte Carlo solver is proposed. 
   
   \end{itemize}

The proposed digital twin was illustrated by using a single-degree of freedom system with mass evolution and stiffness evolution, both individually and jointly. Both mass and stiffness evolution was considered to be of multi-scale nature. The key observations are summarized below: \begin{itemize}

    \item When equally spaced sensor data (clean)    over the service life of the physical system is available, the proposed ME-GP based digital twin reduces to a simple GP based digital twin.     This indicates that a single GP expert is able to describe the time-evolution of the system parameters.

    \item The importance of collecting data over a longer time-window is showcased in this paper.    As the observation time-window increases, the digital twin predictions become more accurate.

    \item The importance of collecting more data is also illustrated in this paper. We showed the digital twin predictions to improve as more sensor data was provided to the framework.

    \item It is important to collect cleaner data from the physical system. This is because as the noise in the data increases, the digital twin seems to deviate from the true solution.   Some denoising techniques can be useful in this regards.

    \item The proposed approach being (partially) Bayesian in nature provides predictive uncertainty.    For most of the results presented in this study, the true solution was enveloped by the predictive uncertainty, indicating that the uncertainty due to noisy and limited data is properly captured.    However, there are a few results where the predictive uncertainty was not properly captured.    This can probably be avoided by employing a completely Bayesian framework where all the parameters are treated in a probabilistic sense.    
    
    \end{itemize}

\section{Conclusions}\label{sec:conclusions}

Digital twins of dynamical systems encountered in engineering and technology can benefit from the use of multiple time-scales. The solution of such systems is difficult from a computational point-of-view as we need a time-step of the order of the fastest scale. As a result, tasks such as health-monitoring, damage prognosis and remaining useful-life prediction becomes excessively difficult. To address this issue, we present a machine learning-based digital twin framework for multi-timescale dynamical systems. The proposed digital twin has two major components: (a) a physics-driven nominal model (generally represented by ordinary or partial differential equation(s)) and (b) a data-driven machine learning model. We use the physics-driven nominal model for data processing and predictions and the machine learning model for learning the time-evolution of the system parameters. We propose to use a mixture of experts (MOE) as the machine learning model of choice. As an expert within the MOE framework, we propose to use GP. The basic idea is that each of the experts will track the temporal evolution of the system parameters at a single scale. For learning the hyperparameters of the proposed model, an algorithm based on expectation-maximization and sequential Monte Carlo sampler is proposed. The proposed training algorithm is of hybrid nature where some of the parameters are treated in a Bayesian sense while the point-estimates for others are provided.

The results obtained using the proposed approach show its ability in predicting the time-evolution of the system parameters, even outside the observation time-window. However, extremely sparse and highly noisy data can affect the performance of the proposed framework. Moreover, collecting data over a longer observation window can drastically improve the performance of the proposed framework. The proposed framework being partially Bayesian quantifies the uncertainty due to limited and noisy data. In most of the cases, the predictive uncertainty is found to envelop the true solution, indicating that the uncertainty is properly captured. However, for some cases, the true solution is found to be outside the envelope. This is probably because of overfitting and employing a fully-Bayesian framework can help us overcome this issue.

Despite the several interesting findings of the current work, there are two possible extensions that need to be pursued in future. First, the illustration has been carried out using a single degree of freedom system. While this helps us in understanding the functionality of the proposed framework, further investigation on multi-degree-of-freedom (MDOF) systems is necessary. For MDOF systems, one major challenge will be in the data-processing step. This is because deriving closed-form relation between frequency measurements and system parameters is difficult, if not impossible. Secondly, we have used a damped natural frequency to be our observation. However, this is a derived quantity based on strain measurements. It is necessary to extend the framework to directly learn the evolution of mass and stiffness from time-history measurements.

\section*{Acknowledgements}

SA  acknowledges the financial support from The Engineering Physical Science Research Council (EPSRC) through a programme grant EP/R006768/1.

\end{document}